\documentclass{osa-article}

\journal{oe}

\articletype{Research Article}

\usepackage{lineno}
\usepackage{gensymb} 
\usepackage{verbatim} 
\usepackage{color} 
\usepackage{xfrac}

\usepackage{caption}
\usepackage{subcaption}
\usepackage{mathtools} 
\usepackage{amsmath}
\usepackage{multirow}
\usepackage{cleveref}

\begin{document}

\title{Shot Noise Reduction in Radiographic and Tomographic Multi-Channel Imaging with Self-Supervised Deep Learning}

\author{Yaroslav Zharov\authormark{1,5,*,$\dagger$}, 
Evelina Ametova\authormark{1,2,$\dagger$}, 
Rebecca Spiecker\authormark{1}, 
Tilo Baumbach\authormark{1,3}, 
Genoveva Burca\authormark{6,4,2},
Vincent Heuveline\authormark{5}
}

\address{\authormark{1}Laboratory for Applications of Synchrotron Radiation (LAS), Karlsruhe Institute of Technology, Germany\\
\authormark{2}Department of Mathematics, The University of Manchester, United Kingdom \\
\authormark{3}Institute for Photon Science and Synchrotron Radiation (IPS), Karlsruhe Institute of Technology, Germany\\
\authormark{4}ISIS Pulsed Neutron and Muon Source, STFC, UKRI, Rutherford Appleton Laboratory, UK\\
\authormark{5}Engineering Mathematics and Computing Lab (EMCL), Interdisciplinary Center for Scientific Computing (IWR), Heidelberg University, Germany\\
\authormark{6}Diamond Light Source, Harwell Campus, Didcot, Oxfordshire, OX11 0QX, UK\\
\authormark{$\dagger$}Contributed equally}

\email{\authormark{*}yaroslav.zharov@kit.edu} 



\begin{abstract}

Noise is an important issue for radiographic and tomographic imaging techniques. It becomes particularly critical in applications where additional constraints force a strong reduction of the Signal-to-Noise Ratio (SNR) per image. These constraints may result from limitations on the maximum available flux or permissible dose and the associated restriction on exposure time. 
Often, a high SNR per image is traded for the ability to distribute a given total exposure capacity per pixel over multiple channels, thus obtaining additional information about the object by the same total exposure time. These can be energy channels in the case of spectroscopic imaging or time channels in the case of time-resolved imaging.    
Conventional image denoising methods work on a per-image basis and rely on certain assumptions concerning image properties. Consequently, they perform well when the assumptions are met and fail otherwise. At the same time, tremendous progress in machine learning demonstrated that data-driven methods are much more flexible in accommodating various image characteristics.

In this paper, we report on a method for improving the quality of noisy multi-channel (time or energy-resolved) imaging datasets. The method relies on the recent \emph{Noise2Noise} (N2N) \cite{Lehtinen2018} self-supervised denoising approach that learns to predict a noise-free signal without access to noise-free data. N2N in turn requires drawing pairs of samples from a data distribution sharing identical signals while being exposed to different samples of random noise. 
The method is applicable if adjacent channels share enough information to provide images with similar enough information but independent noise. 
We demonstrate several representative case studies, namely spectroscopic (k-edge) X-ray tomography, \textit{in vivo} X-ray cine-radiography, and energy-dispersive (Bragg edge) neutron tomography. In all cases, the N2N method shows dramatic improvement and outperforms conventional denoising methods. For such imaging techniques, the method can therefore significantly improve image quality, or maintain image quality with further reduced exposure time per image.
\end{abstract}

\section{Introduction}

Many imaging modalities rely on the penetration ability of radiation to make the interior of an object visible. Physical interactions of radiation and matter, such as absorption, scattering, or phase shifts, can be used to obtain contrast inside the object of interest. Commonly, either \emph{radiography} (single view projection) or \emph{tomography} (multiple views with subsequent volumetric image reconstruction) are acquired. As particle or photon emission and detection are stochastic processes, and often source flux and detector efficiency are limited, longer exposure time improves image quality. However, there are many scenarios where sufficient exposure can not be achieved. 
An obvious example is \emph{in vivo} imaging, where the radiation dose ultimately limits the amount of information acquired~\cite{Moosmann2013}. Another example is spectroscopic imaging with a polychromatic beam, where the detected intensity or particle counts are distributed across multiple energy bins.
This leads to a significant noise per energy channel or requires a dramatic increase of exposure times, hence, limiting the experiment throughput~\cite{warr2021enhanced}. Both imaging modes can be generalized as \emph{multi-channel} images.  The paper addresses a number of cases when the channels of multi-channel images share a sufficient amount of common structural information but are affected by independent noise samples.

Given the aforementioned physical constraints, we often need to rely on image processing techniques to improve image quality and extract valuable data. A group of methods for improving image quality that is affected by noise is called \emph{denoising}. Similar to other domains, methods based on \emph{Machine Learning} (ML) have revolutionized denoising \cite{Ilesanmi2021}. In this paper, we demonstrate an ML approach to improve the quality of underexposed images in challenging applications such as spectroscopic k-edge X-ray tomography, \textit{in vivo} X-ray radiography, and energy-dispersive Bragg-edge neutron tomography. The method is based on the recent Noise2Noise (N2N) self-supervised denoising approach~\cite{Lehtinen2018}. 
The main assumption enabling the N2N method is formulated as follows:
Consider two images $I_1$ and $I_2$ that share the same structural information $S$ but are affected by independent and identically distributed (iid) instances of noise $\sigma_1$ and $\sigma_2$. 
If the model is trained to predict $I_1$, given $I_2$ as input, the best prediction possible is $S$, because $\sigma_2$ is conditionally independent of $\sigma_1$, given $S$.

This paper is organized as follows. First, we outline the related work to put our work in context. Then, the N2N method is described, followed by three rigorous case studies. Finally, a discussion of our findings in the context of multi-channel imaging is provided.

\section{Related Work}

Artifacts are inherent to digitally acquired images, as an acquisition function is subject to many uncertainties and in general, is not accurately known. The acquisition function includes source or detector heterogeneity,  optical distortion by optical elements or diffraction during wave-field propagation, and inherent noise driven by the stochastic nature of particle emission, detection, and interactions. As optical distortion is a misplacement of information, a distortion map can be estimated and applied to compensate for it. For a number of cases, simple flat and dark field corrections are applicable. Flat and dark fields refer to the measurement of detector response with and without source illumination, respectively. Finally, denoising is used to compensate for the inherent stochastic noise. Hence, the denoising problem is to restore the (deterministic) signal $S$ from a noisy observation $I$~\cite{Gonzalez2008}:

\begin{equation}
    I = S + \sigma(S)
\end{equation}

\noindent where $\sigma(S)$ is the inherent noise of the imaging device. This noise depends not only on the stochasticity of the particles (neutrons, photons, etc.) and the electronics but additionally on the distortions, and also on the transformation applied to correct the image. All this makes a closed-form distribution estimation problematic.

The existing image denoising approaches can be roughly categorized into two large groups: classical image processing and ML approaches. Typically, classical image processing approaches work in a single-image manner and incorporate expert beliefs about the nature of noise. ML approaches, on the other hand, employ the idea of fitting a data-driven model entirely without or with minimal expert knowledge about the nature of the data.

\subsection{Classical Image Processing}

The basic spatial filtering methods are mean, median, or Gaussian kernel filters~\cite{Gonzalez2008}. For each pixel, these filters select a new value, based on the weighted values of the neighboring pixels. These filters are fast, robust, well-understood, and work fairly well in many situations. The main drawback of these classical filters is their tendency to blur sharp edges.

More advanced spatial filtering approaches, e.g., non-local means (NLM), use more information from the whole image \cite{Buades2005}. Instead of taking an average of the direct neighbourhood of the pixel, NLM takes an average of the large region, weighted by the similarity between the ``donor'' and the ``recipient'' pixel. The process of revisiting multiple locations in the image, comparing their surroundings, and computing the average can take minutes for one image. In return, this method is capable of producing sharper denoised images~\cite{Fan2019}.

Alternatively, denoising can be formulated as an optimization problem and regularization can be used to incorporate some prior knowledge about the image properties~\cite{gu2019brief}. These methods are very powerful but require deep mathematical knowledge and handcrafted regularizers in many cases, making their application for experimental data challenging. One of the most successful regularizers is Total Variation (TV) which encourages piece-wise constant image regions with sharp boundaries~\cite{rodriguez2013total}.
In summary, classical methods require fine tuning of parameters by an expert to balance smoothing and denoising. Hence there is a significant risk of information loss if applied incorrectly. A more comprehensive overview of classical denoising methods can be found elsewhere~\cite{Fan2019}.

\subsection{Machine Learning Approaches}

The evolution of classical methods may be seen as a series of steps taken to increase the amount of information used to correct a single pixel value. In this respect, ML-based approaches appear as a natural further step: a \emph{model}, trained to correct the noise, implicitly incorporates knowledge about the whole dataset.

Early ML-based image denoising approaches worked in a supervised manner, i.e. a model was trained on a set of noisy images to predict a noise-free image (\emph{target}). Recently, authors of the N2N method demonstrated that there is no need for a noise-free target: if one uses a pair of noisy images (affected by iid instances of the noise) as an input and as a target for the training, the model will predict the noise-free image~\cite{Lehtinen2018}. The underlying intuition is that independent instances of noise are uncorrelated and cannot be predicted, hence the model is forced to extract features. Even though N2N does not explicitly require a set of noise-free images, the authors of ~\cite{Lehtinen2018} synthetically formed noisy pairs by adding noise to noise-free images.

There have been several attempts to extend the N2N method for denoising problems where pairs of images are not naturally available. Noise2Self~\cite{Batson2019} and Noise2Void~\cite{Krull2019} generate the required pair of images by taking random pixels in the noisy image and disturbing them with yet another noise distribution. In this way, multiple training pairs can be constructed from a single noisy image. Noise2Stack~\cite{Papkov2021} was designed for three-dimensional tomographic data and is based on an assumption that tomographic data is typically smooth. Therefore, slice-to-slice changes are assumed to be significantly smaller than the slice-wise variability caused by noise, hence, neighbouring slices can be used for training.

Alternatively, constrained autoencoders can be used to denoise images~\cite{Vincent2008}. During the training, autoencoders use the same image both as input and target and attempt to compress (encode) the input image into its lower-dimensional representation. The denoising properties of this approach rely on the assumption that the noise, due to its stochastic nature, is harder to encode, than the signal. To additionally limit the capacity of the model to store information about the noise, it can be restricted by limiting the computational capacity of the model, lowering the dimensionality of the learned representation, or introducing synthetic noise into it~\cite{Vincent2008}. However, the autoencoders are inefficient if the noise is spatially correlated and can be easily memorized by the model. The Hierarchical DivNoising~(HDN) method addresses this issue by training a variational autoencoder with a noise model imposed over output~\cite{Prakash2022}. The authors proposed a way to find particular components of the model that encode information about the noise so that they can remove those components. Even though the proposed methods provide a valuable alternative to the N2N approach, the authors highlight that the N2N approach is a hard-to-beat baseline~\cite{Prakash2022}.

\section{Model Training}

The N2N method assumes that a pair of images contains the same signal and iid noise. Our adaptation of the method to multi-channel image data takes its inspiration from the denoising of Synthetic Aperture Radar (SAR) images~\cite{Dalsasso2022}. In SAR imaging, both the phase and amplitude of received microwaves are measured in each pixel; commonly the phase information is ignored. However, the authors demonstrated that the amplitude and the phase contain complementary information and can be used as a basis for N2N denoising. We hypothesize that in multi-channel imaging, adjacent time frames or energy levels indeed share sufficiently similar signals, and have noise samples close to being iid. Therefore, we generate the required pair images based on this hypothesis. To help the model catch complex spatial structures of the signal, we also feed it with multiple adjacent energies or time frames as input whenever it does not result in oversmoothing.

Following~\cite{Krull2019,Batson2019,Lehtinen2018}, we use the fully-convolutional neural networks as a model architecture. We employ U-Net with ResNet-50 (as implemented in \cite{PavelIakubovskii2019}) as the backbone and rely on the Adam optimizer with a $3\times10^{-4}$ learning rate, without scheduling. Referring to the model, we will use $f_\theta$ and model interchangeably, where $\theta$ denotes trainable parameters of the neural network. The training, therefore, is the process of minimization of the proposed loss function (defined for specific experiments) by changing the parameters $\theta$. We augment each image pair with random crops, shifts, scale, rotations, distortions, and different types of blur. We acknowledge that there is room for quality improvement via larger models, modern architectures, better optimization procedures, or more aggressive augmentations. The sensitivity study of training parameters is a topic of future investigation.

\section{Experiments} 

\subsection{Simulated spectral X-ray tomography}
\label{sec:simulation}

As a first case study we discuss the applicability of N2N to energy-dispersive X-ray tomography, which is of interest for biomedical imaging~\cite{warr2021enhanced}. The polychromatic emission of laboratory X-ray tube sources is suitable to provide sufficient photon flux. However, the broadband spectrum also leads to disadvantages in quantitative analysis. In the conventional absorption mode, each detector pixel integrates all photons irrespective of their energy. Since attenuation is a function of photon energy, conventional tomographic reconstruction might exhibit so-called beam-hardening artifacts~\cite{davis2008modelling}. 
However, acquisition with an energy-dispersive X-ray detector allows segmenting materials that can be inseparable in polychromatic absorption contrast. These are materials with similar mean polychromatic absorption, but with spectrum showing sharp discontinuities at energies equal to the binding energies of the core-electron states, so-called \emph{absorption-edges (K, L, M) edges}. Thie energy spectrum in each reconstructed voxel can be used to identify the corresponding material. Highly energy-dispersive (so-called hyperspectral) X-ray detectors have yet a limited total pixel number but an energy resolution of about 1~keV~\cite{egan20153d}, allowing to distinguish even neighboring chemical elements. However, a high spectral energy resolution entails long exposure times since the acquired counts are distributed over multiple bins. Therefore, a state-of-the-art reliable denoising approach might help to improve the experimental throughput. In this study, to ensure strictly controlled conditions, we simulated the tomographic acquisition.

\subsubsection{Data}

We generated a volumetric phantom by combining several three-dimensional point clouds: two Swiss rolls, two moon crescents, and an s-curve. All point clouds were generated by the Scikit-Learn library~\cite{Pedregosa2012}; to convert 2D point clouds to 3D, the third axis was added by randomly sampling from the Uniform distribution. To convert the point clouds to a raster volume, we selected the size (in voxels) of each point.
To resolve the ambiguous cases (when several materials appeared in the same voxel), we selected the priority order 
 and always assigned the material with the highest order to fill the ambiguous voxel. The spatial size of the phantom was set to $512 \times 512 \times 512$. In a rough structure, all slices of the dataset are the same. However, the surface texture varied because of the random nature of the point clouds. A single slice is shown in Figure \ref{fig:simulation_qualitative:40} (left).

We assigned the simulated objects with energy-dependent mass attenuation coefficient (MAC) of Europium ($\prescript{}{63}{\mathbf{Eu}}$, k-edge = 48.5~keV), Gadolinium ($\prescript{}{64}{\mathbf{Gd}}$, k-edge = 50.2~keV), Ytterbium ($\prescript{}{70}{\mathbf{Yb}}$, k-edge = 61.3~keV), Lutetium ($\prescript{}{71}{\mathbf{Lu}}$, k-edge = 63.3~keV), and Uranium ($\prescript{}{92}{\mathbf{U}}$, k-edge = 115.6~keV). The background was assigned with MAC of air. This particular choice of materials was inspired by the study of the separability of k-edge nanoparticles presented in~\cite{getzin2018increased}. Two pairs of materials have neighbouring atomic numbers, hence very close k-edges, and are barely distinguishable in a noisy image; Uranium was added to have a k-edge in the noisiest part of the spectrum, to check the ability of the method to locate the k-edge in extreme noise conditions. 

\begin{figure}
     \centering
         \includegraphics[width=\textwidth]{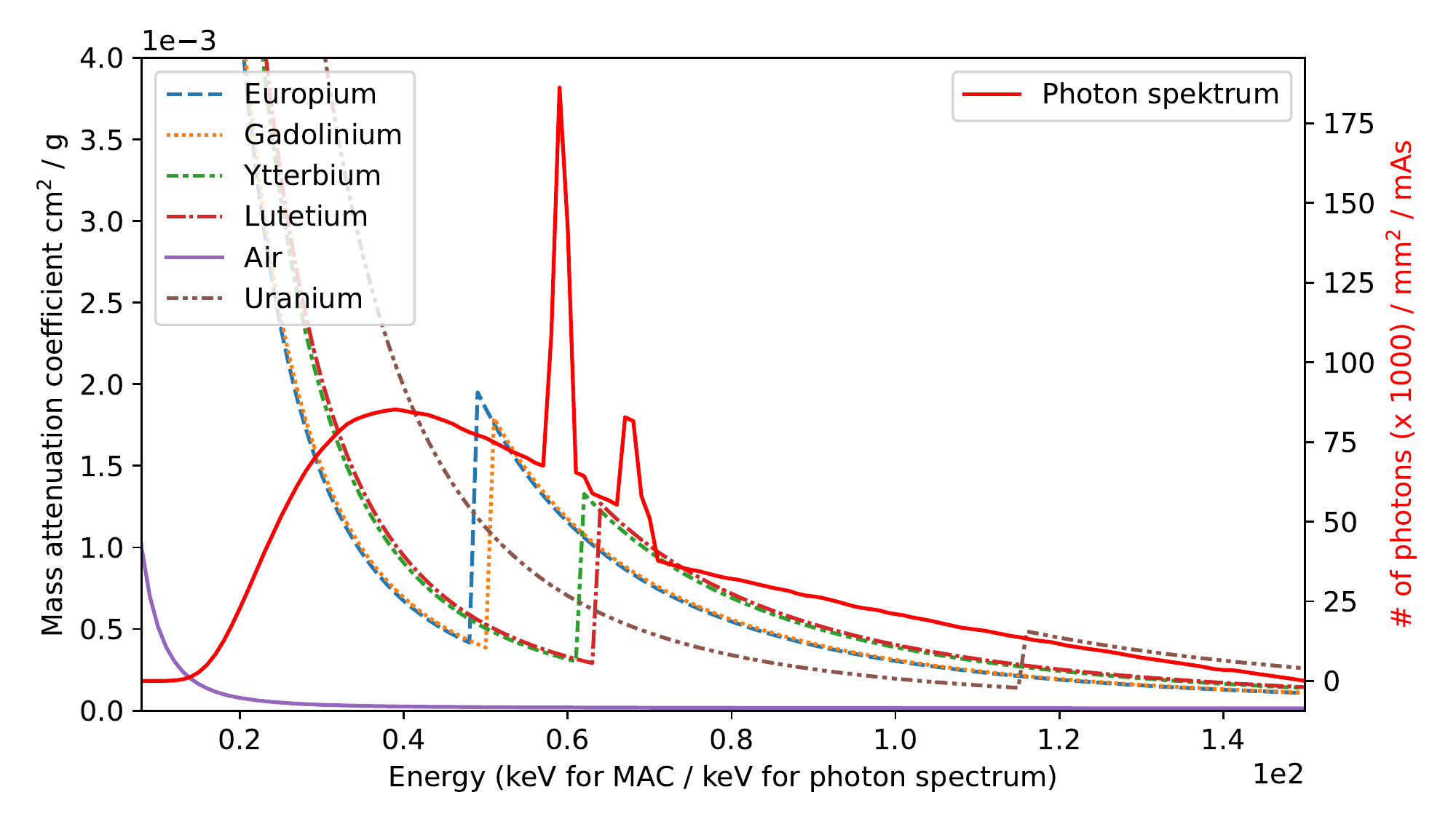}
        \caption{\label{fig:spectra} The materials and source characteristics used to simulate spectral X-ray tomography. For materials, energy-dependent mass attenuation coefficients (MAC) are presented, and for the simulated Boone/Fewell source, we present the source profile. We selected two pairs of materials with close k-edges that are hard to resolve and one material with the k-edge in the low-flux zone of the source.}
\end{figure}

We used the MATLAB package \texttt{PhotonAttenuation} to generate the energy-dependent MAC of the selected materials~\cite{tuszynski2006}. A spectrum profile of a Boone/Fewell source with the tube potential 150~kV (no kV Ripple and filters) was generated using the MATLAB package \texttt{spektr}~3.0~\cite{punnoose2016spektr}. The obtained source spectrum was normalized and scaled to have a maximum value of $175 \times 10^3$ photons / $\mathrm{mm^2}$ to imitate short exposure acquisition. MAC of selected materials and the source spectrum are shown in Figure~\ref{fig:spectra}.

We generated 135 energy bins between 15 and 150~keV with a 1~keV step. For each bin, we simulated 120 equally-spaced parallel-beam CT projections over 180\degree. The spectral characteristics of the material and the source are shown in Figure~\ref{fig:spectra}. We used the conventional FBP algorithm to reconstruct tomographic data (implemented in in~\cite{jorgensen2021core}). Examples of reconstructed slices for 40~keV (high flux) and 140~keV (low flux) are shown in Figure~\ref{fig:simulation_qualitative:140} (right). As expected, at 140~keV the reconstructed slice is uninterpretable.

\subsubsection{Training and Processing}

The model $f_{\theta}$ was trained by optimizing 

\begin{equation}
    \mathbb{E}_{i,j} \Vert f_{\theta}(x_{i,j-1}, x_{i,j+1}) - x_{i,j} \Vert_1 \xrightarrow[\theta]{} \min, 
\end{equation}

\noindent where $x_{i,j}$ is a projection acquired at the transmission angle $i$ and in the energy bin $j$. We randomly split the whole set of projection angles into a training set and a validation set with a ratio of $80/20$. We do not select a test set, since in our experiments, we do want to overfit for the exact dataset and do not require generalization. Note that the energy level $j$, which is required to be predicted by a model, should not be fed into the model to avoid the trivial solution.
Only adjacent $j-1, j+1$ levels should be used. This forms a gap of one energy level in the inputs. 

During the inference, we feed the model with the adjacent energy bins without the gap used in training, to avoid blur in the spectral domain: 

\begin{equation}
    \tilde{x}_{i,j-0.5} = f_{\theta}(x_{i,j-1}, x_{i,j}).
\end{equation}

However, since the model predicts the energy level which is averaged between two input levels, it will inevitably predict an energy level between two adjacent ones used as input. It is important yet easy to compensate for this.

As before, the denoised tomographic datasets were reconstructed with the conventional FBP algorithm. Here, each energy bin was reconstructed separately resulting in 135 volumes. To obtain the spatial distribution of individual materials in the sample, we performed material decomposition as described in~\cite{ametova2021crystalline}. The employed decomposition relies on the assumption that each voxel is a unit volume and each material occupies a volume fraction in this unit volume (the fraction can be 0). Under this assumption, a voxel-wise sum of all material maps is equal to 1 in each voxel.

\subsubsection{Results}

\begin{figure}
    \begin{subfigure}{.5\textwidth}
      \begin{subfigure}{\textwidth}
          \centering
          \includegraphics[width=\linewidth]{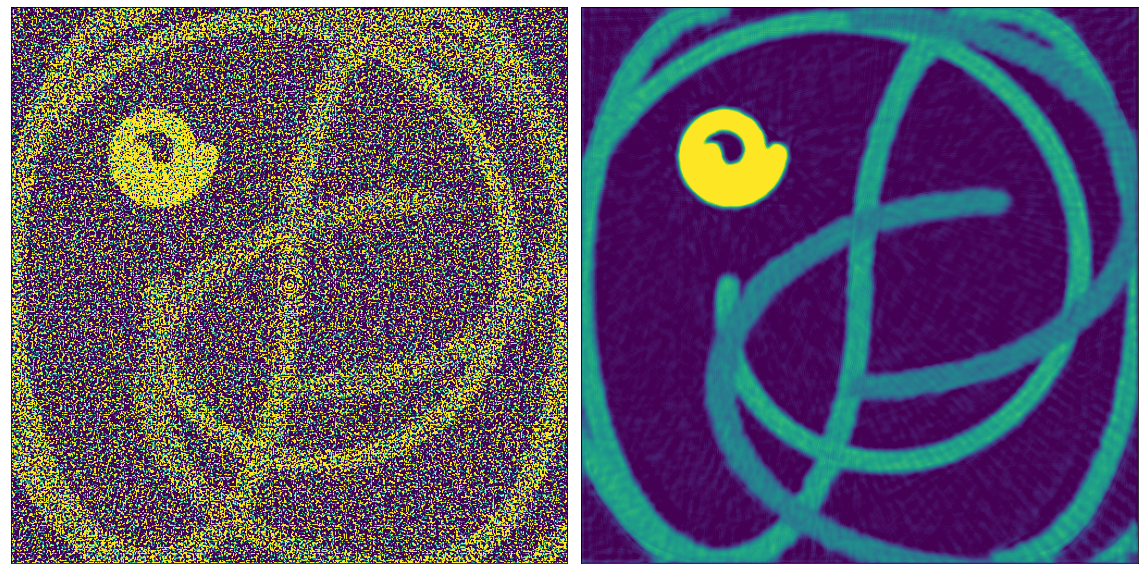}  
          \caption{Noisy and denoised slices at 40 keV (near peak source flux)}
          \label{fig:simulation_qualitative:40}
        \end{subfigure}
         \newline
        \begin{subfigure}{\textwidth}
          \centering
          \includegraphics[width=\linewidth]{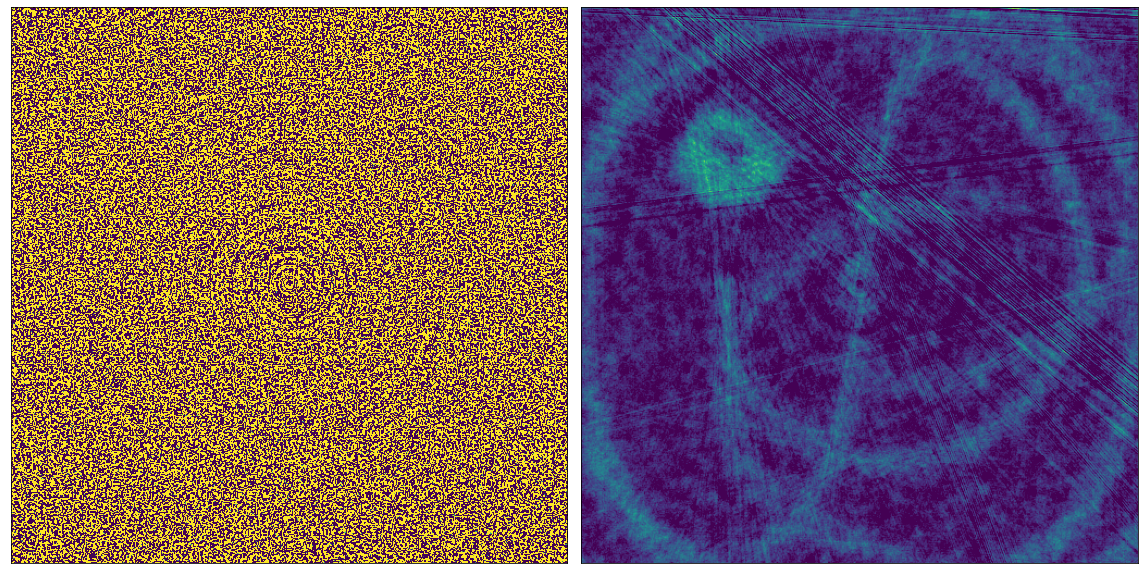}  
          \caption{Noisy and denoised slices at 140 keV (low source flux)}
          \label{fig:simulation_qualitative:140}
        \end{subfigure}
    \end{subfigure}
    \begin{subfigure}{.5\textwidth}
      \centering
      \includegraphics[width=\linewidth]{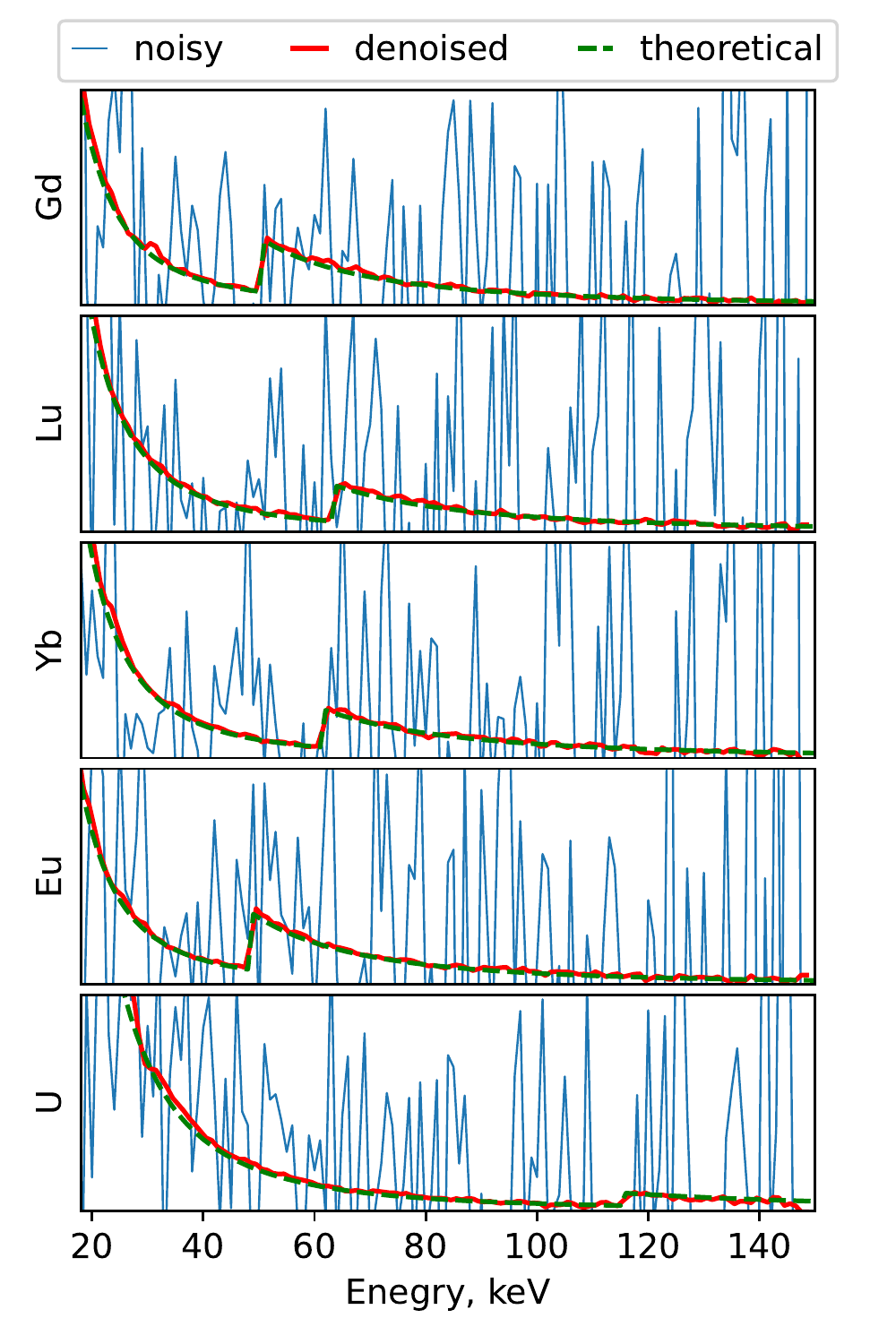}  
      \caption{Spectral profiles}
      \label{fig:simulation_qualitative:spectra}
    \end{subfigure}
    \caption{Qualitative examination of the denoising of the simulated energy-resolved X-ray CT of a specially devised phantom. On the left, present a noisy (left) and denoised (right) transverse slice both near peak (top) and low (bottom) flux. On the right, we present a comparison of theoretical, noisy, and denoised spectra for different materials. For each material, we selected one representative pixel. Note how denoising is able to recover information even in extremely noisy cases both spatially (b) and spectrally (see the slight k-edge of the Uranium on the (c) plot).}
    \label{fig:simulation_qualitative}
\end{figure}

\Cref{fig:simulation_qualitative:40,fig:simulation_qualitative:140} show two-dimensional slices for selected (individual) energy bins. N2N demonstrates the drastic quality improvement of the reconstruction. For 40~keV (high source flux, \Cref{fig:simulation_qualitative:40}) the reconstructed slice appears to be almost noise-free; the slice shows sharply defined objects where all original structures become clearly visualized. Although no signal seems to be visible in the 140~keV slice (\Cref{fig:simulation_qualitative:140}) prior to denoising, N2N is able to partially recover the structures in the slice.

Single noisy energy spectra are used for one voxel per material component and denoised spectra are reconstructed for the voxels and plotted in Figure~\ref{fig:simulation_qualitative:spectra} along with the theoretical MAC. The voxel positions within the materials were chosen arbitrarily. Noise reduction results in sharp and accurately positioned k-edges, aiding further material decomposition. Even the slight uranium k-edge is visible in the denoised spectrum.

\begin{figure}
    \begin{subfigure}{.5\textwidth}
      \centering
      \includegraphics[width=\textwidth]{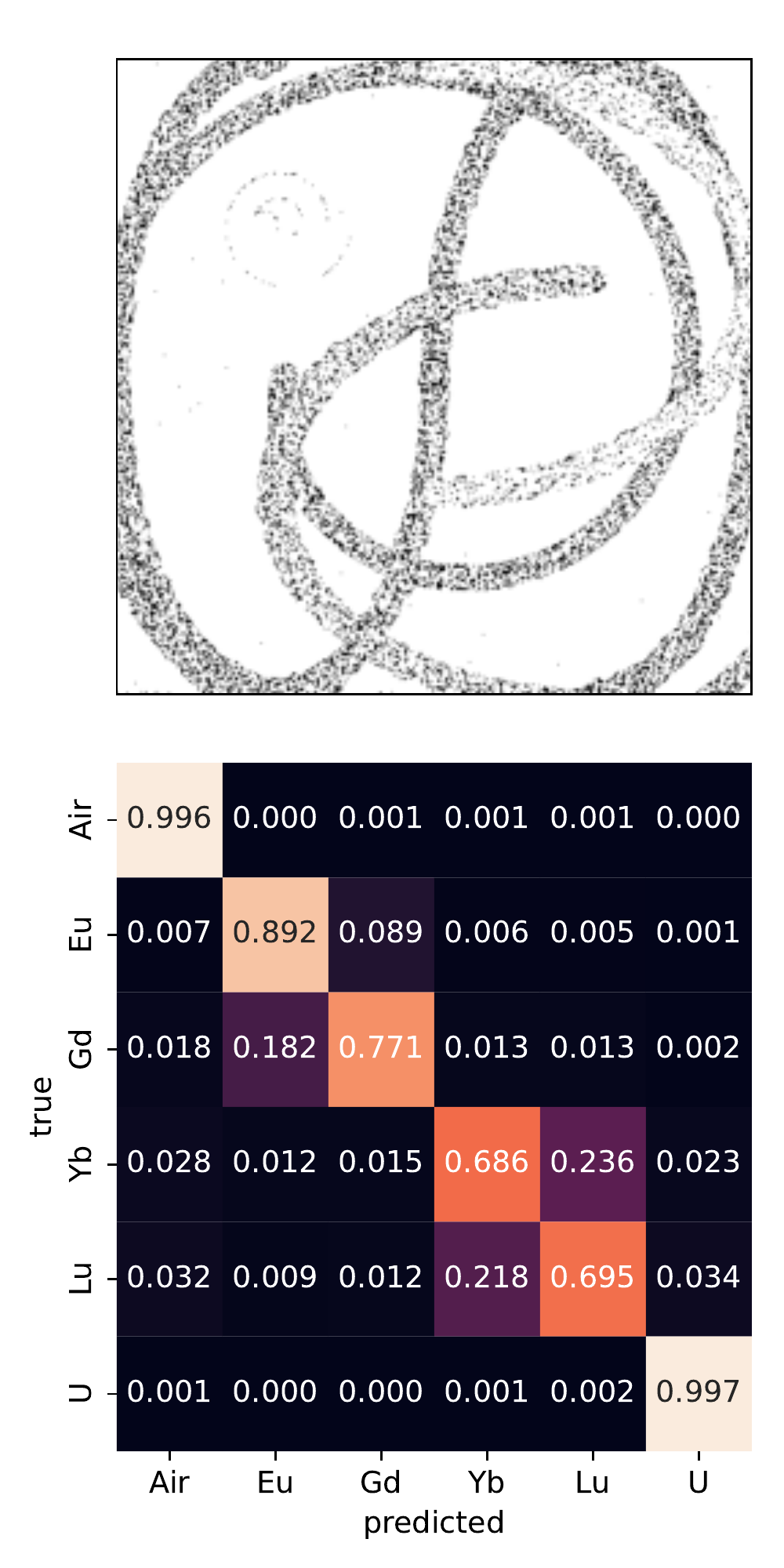}  
      \caption{Noisy}
      \label{fig:simulation_analysis:low_exposure}
    \end{subfigure}
    \begin{subfigure}{.5\textwidth}
      \centering
      \includegraphics[width=\textwidth]{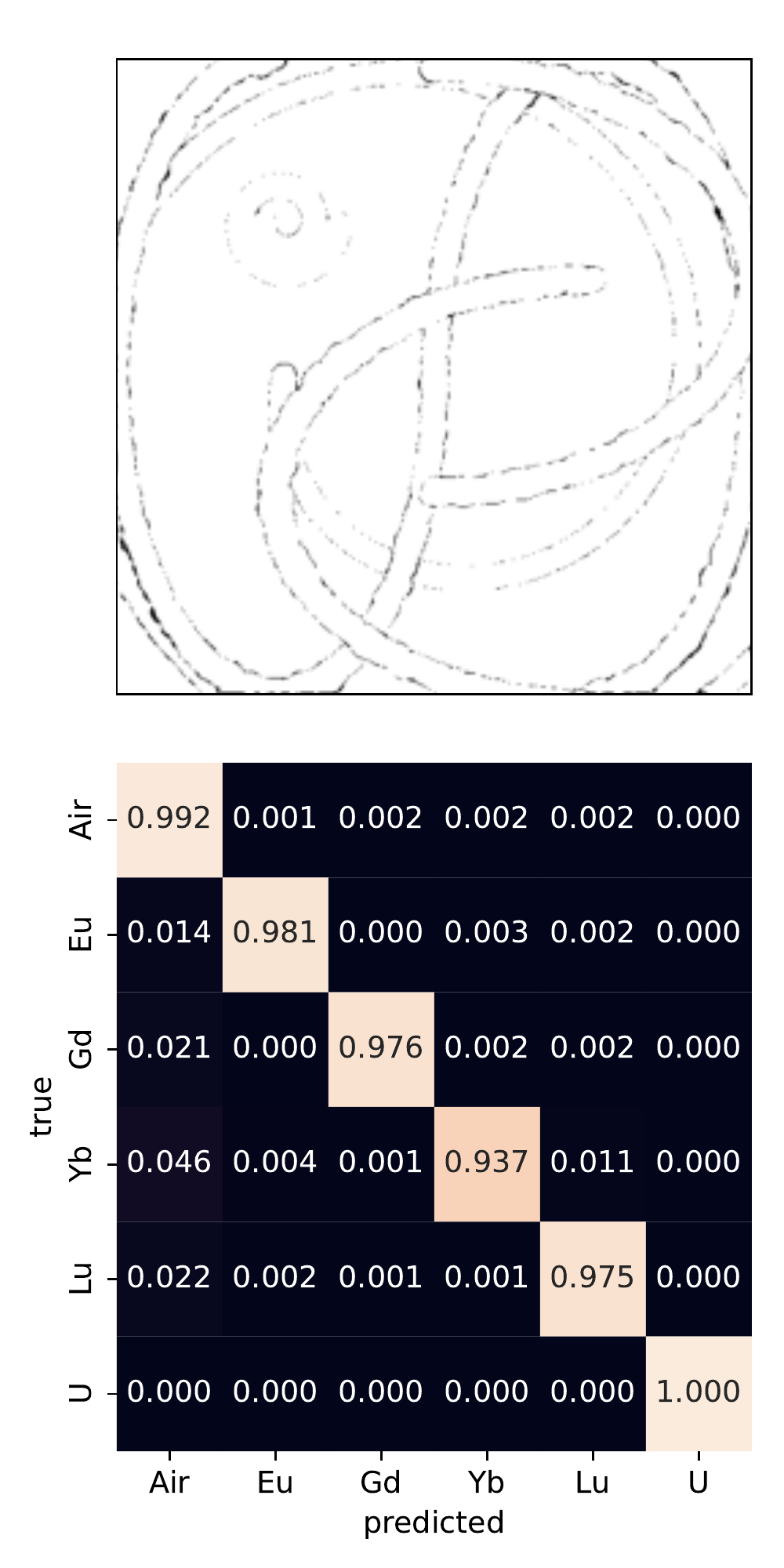}  
      \caption{Denoised}
      \label{fig:simulation_analysis:low_exposure_denoised}
    \end{subfigure}
    \caption{Quantitative examination of the denoising of the simulated energy-resolved X-ray CT of a specially devised phantom.
    We study the quality of denoising through the lens of further material decomposition.
    On the top row, we present the binarized material decomposition error. Pixels that are black were assigned the wrong material.
    On the bottom row, we present the confusion matrix of the material decomposition for different materials, where ideal decomposition should yield the identity matrix.
    We note that materials with close k-edges are frequently confused before the denoising, and after the denoising, the confusion mainly comes from spatial smoothing.} 
    \label{fig:simulation_analysis}
\end{figure}

To quantitatively evaluate denoising results, we perform the material decomposition. Since the sum of volume fractions corresponding to each material, obtained through material decomposition, is bound to 1 in each voxel, we can treat the estimated volume fractions as probabilities. Hence, the task of material decomposition can be considered a classification problem and the related quality assessment metrics can be applied to quantitatively assess the results. The comparison results are shown in the Figures~\ref{fig:simulation_analysis:low_exposure} and~\ref{fig:simulation_analysis:low_exposure_denoised}. In the top row we show the binarized material decomposition error (black corresponds to erroneous material prediction). The confusion matrices between the predicted and true materials for each pixel are presented in the bottom row (perfect classification results in the identity confusion matrix). A high level of noise in the simulated data causes misclassifications between close materials (e.g., Lutetium and Ytterbium). Also, as it is visible on the top row, these errors are distributed evenly throughout the sample. Hypothetically, this can be compensated by enforcing an assumption of material homogeneity. However, this assumption might cause severe errors close to material interfaces.  The errors in the denoised volume are mostly concentrated around the borders (see the top row), and mainly correspond to misclassification for air due to slight blur (see the bottom row). But overall, the confusion matrix for the denoised dataset is considerably closer to the identity matrix. 

We also present the Area Under Precision-Recall Curve (AUPRC), measured for each material~\Cref{tab:simulation_auprc}. AUPRC for ideal classification is 1. AUPRC results additionally highlight improvement after denoising: N2N provides a boost of more than $10\%$ of mean AUPRC for the downstream material decomposition. To assess quality loss caused by reconstruction itself (without any effect of denoising), we generated another set of projections with very high flux (all other parameters remained constant). Material decomposition for this volume shows a mean AUPRC of $0.999$, with the lowest \emph{precision} of $0.996$ for the air. We conclude that the reconstruction losses are negligible in this experiment.

\begin{table}
    \centering
   \begin{tabular}{cr||c|c|c|c|c|c|c}
                & & Air & Eu & Gd & Yb & Lu & U & mean\\
                \hline
                \hline
                \multirow{2}{*}{\rotatebox[origin=c]{90}{AUPRC}}
                & noisy & 0.999 & 0.917 & 0.873 & 0.787 & 0.645 & 0.998 & 0.870 \\
                & denoised & 0.999 & 0.998 & 0.998 & 0.996 & 0.995 & 0.999 & 0.998
        \end{tabular}
        \caption{Quantitative examination of the denoising of the simulated energy-resolved X-ray CT of a specially devised phantom. We numerically compare the material decomposition quality before and after denoising. The denoising provides a prominent quality boost for material decomposition.}
        \label{tab:simulation_auprc}
\end{table}

\subsection{Neutron imaging}
\label{sec:neutron}

As a second case study we discuss the applicability of N2N to energy-dispersive Bragg edge neutron tomography. Neutron imaging provides a complementary contrast to conventional X-ray imaging. Neutrons mainly interact with atomic nuclei, in this way a neutron beam passing through an object can capture information about the internal material structure. The energy spectrum of the neutron transmission of a polychromatic thermalized neutron beam passing a predominantly polycrystalline material contains sudden and sharp edges at wavelengths equal twice the interplanar distance between scattering planes in dependence of the crystalline properties of the sample material~\cite{fundamentals1993nuclear}. Energy dispersive images can efficiently be acquired by combining a pulsed neutron spallation source and a suitable time-sensitive detector by using the time-of-flight (ToF) method, which employs the energy-dependent neutron velocity for spectral information (the more energetic neutrons, by having higher velocities, reach the detector earlier than less energetic (slower) neutrons). Measuring the time of arrival of the neutrons at the detector and knowing the flight path length, their energies, and the corresponding wavelengths can be determined. 
For TOF methods, high energy resolution requires long flight distances and many time bins in the detector. Hence, only a few pulses per second can be measured, and acquired counts are shared between multiple bins~\cite{Santisteban2001}. 
More details on this acquisition mode can be found elsewhere, for both the measurement setup~\cite{kockelmann2007energy} and applications~\cite{santisteban2002engineering,strobl2009advances}. Neutron facilities are expensive and demand for neutron beamtime exceeds the supply capacity~\cite{Bentley2020}. Therefore, there is a high interest in efficient image denoising techniques to reduce exposure time and subsequently increase experiment throughput.

\subsubsection{Data}
\label{sec:neutron:data}

In this study, we employ the dataset~\cite{jorgensen2019neutron} acquired at the Imaging and Materials Science \& Engineering (IMAT) beamline operating at the ISIS spallation neutron source (Rutherford Appleton Laboratory, U.K.)~\cite{burca2013modelling,kockelmann2018time}. More details on acquisition parameters and preprocessing can be found elsewhere~\cite{ametova2021crystalline}; here we only briefly summarize details relevant to this study. 

A sample contains 6 aluminium tubes: five filled with metallic powder (copper (Cu), aluminium (Al), zinc (Zn), iron (Fe), and nickel (Ni)), and one empty. The neutron detector has $512 \times 512$ pixels, 0.055~mm pixel size. A set of spectral projections were acquired at 120 equally-spaced angular positions over {180\degree} rotation with 15~min exposure. Additionally, 8 flat field images (4 before and 4 after the acquisition) were acquired with the same exposure.

A typical problem of spectral measurements is that noise statistics vary quite drastically across the spectrum. The beam spectrum at the IMAT beamline has a crude bell shape with a peak around 3~{\AA}~\cite{burca2013modelling}. Additionally, the time-sensitive detector suffers from dead time meaning counts loss, hence, additional signal distortions~\cite{tremsin2012high}. To alleviate the count loss problem, the time (wavelength) domain is split in several independent measurement intervals (4 in this case) and a special correction technique is applied to the measured data~\cite{tremsin2012high}. Each interval has an individual bin width; for this study the following bin width was used: $0.7184 \cdot 10^{-3}$~{\AA}, $1.4368 \cdot 10^{-3}$~{\AA}, $\cdot 10^{-3}$~{\AA} and $2.8737 \cdot 10^{-3}$~{\AA}. To benchmark N2N, we generated three additional datasets by rebinning the dataset in the original resolution (2840 energy bins split into 4 measurement intervals with (1141, 814, 424, 464) bins in each). The rebinning was performed individually in each interval by summing every (4, 2, 2, 1), (8, 4, 4, 2), and (16, 8, 8, 4) bins, resulting in datasets with 1366, 681 and 339 wavelength bins, respectively.

\begin{figure}[h!]
\centering
\includegraphics[width=\textwidth]{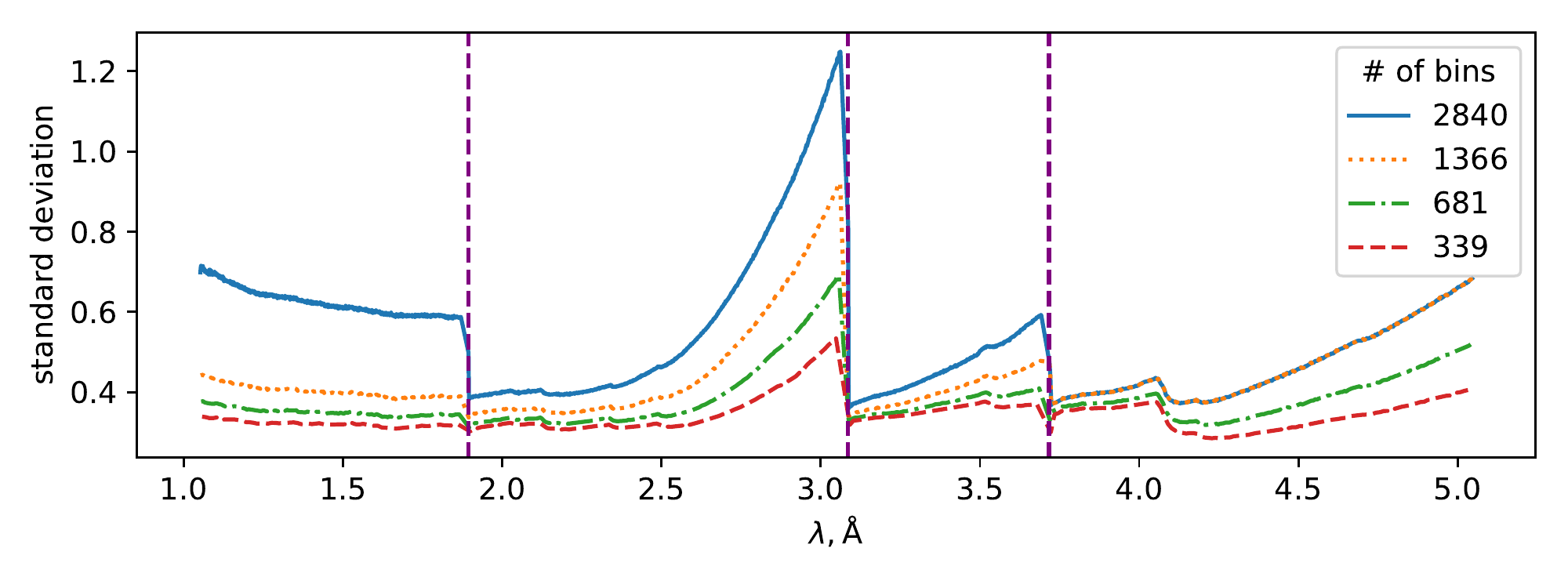}
\caption{Limitations of time-sensitive detectors require splitting the whole wavelength domain into several measurement intervals (4 in our case, brackets depicted with dashed lines). Each interval has an individual wavelength bin width. In this plot, we show how standard deviation (as a proxy characteristic of noisiness) changes with the change in wavelength. Additionally, to benchmark the method, we generated three additional datasets by rebinning the spectral dimension of the original dataset.}
\label{fig:neutron:noise_patterns}
\end{figure}

As a proxy to demonstrate the noisiness of the data, as a function of frequency, we plot the standard deviation of pixel values for one projection angle but different wavelengths along the spectrum in \Cref{fig:neutron:noise_patterns}. Vertical dashed lines separate independent intervals. Note, that standard deviation increases drastically with the increase of flux (the effect of counts loss becomes more apparent).

\subsubsection{Training and Analysis Details}

In this experiment, we compare the effect of noise reduction applied to the projections (N2N(P)) before reconstruction with that of applying it to the already reconstructed slices (N2N(S)). In both cases, we trained a model $f_{\theta}$ by performing essentially the same loss optimization procedure as in the previous case study

\begin{equation}
    \mathbb{E}_{i,j} \Vert f_{\theta}(x_{i,j-1}, x_{i,j+1}) - x_{i,j} \Vert_1 \xrightarrow[\theta]{} \min,
\end{equation}

\noindent where now $x_{i,j}$ can represent either the projection for an angle $i$ and an energy channel $j$, or a reconstructed slice number $i$ and an energy channel $j$. We used $i$ to randomly split the dataset into the training and validation subsets in the $80/20$ ratio.

The combination of the N2N denoising approach and the conventional FBP reconstruction was compared with the advanced iterative reconstruction routine proposed in~\cite{ametova2021crystalline}. The latter relies on expert expectations on how the reconstructed image should look like. Which becomes increasingly complex with increasing complications of the sample under investigation. As in this case the reconstructed samples are expected to appear as solids, \emph{i.e.} homogeneous regions, the authors assumed a piece-wise constant signal in the spatial domain. This prior knowledge is enforced through TV regularization~\cite{rudin1992nonlinear,sidky2006accurate}. The signal in the spectral domain is expected to be piece-wise smooth based on theoretical predictions for the materials employed in this study~\cite{boin2012nxs}. In this case, regularization is achieved through Total Generalized Variation (TGV) prior~\cite{bredies2010total}. Hence, we refer to the iterative reconstruction method as TV-TGV. As before, the reconstruction was implemented in CIL~\cite{Papoutsellis2021}. Code to reproduce results is available from~\cite{evelina_ametova_2021_4884710}.

\subsubsection{Results Discussion}

\begin{figure}
\centering
\includegraphics[width=\textwidth]{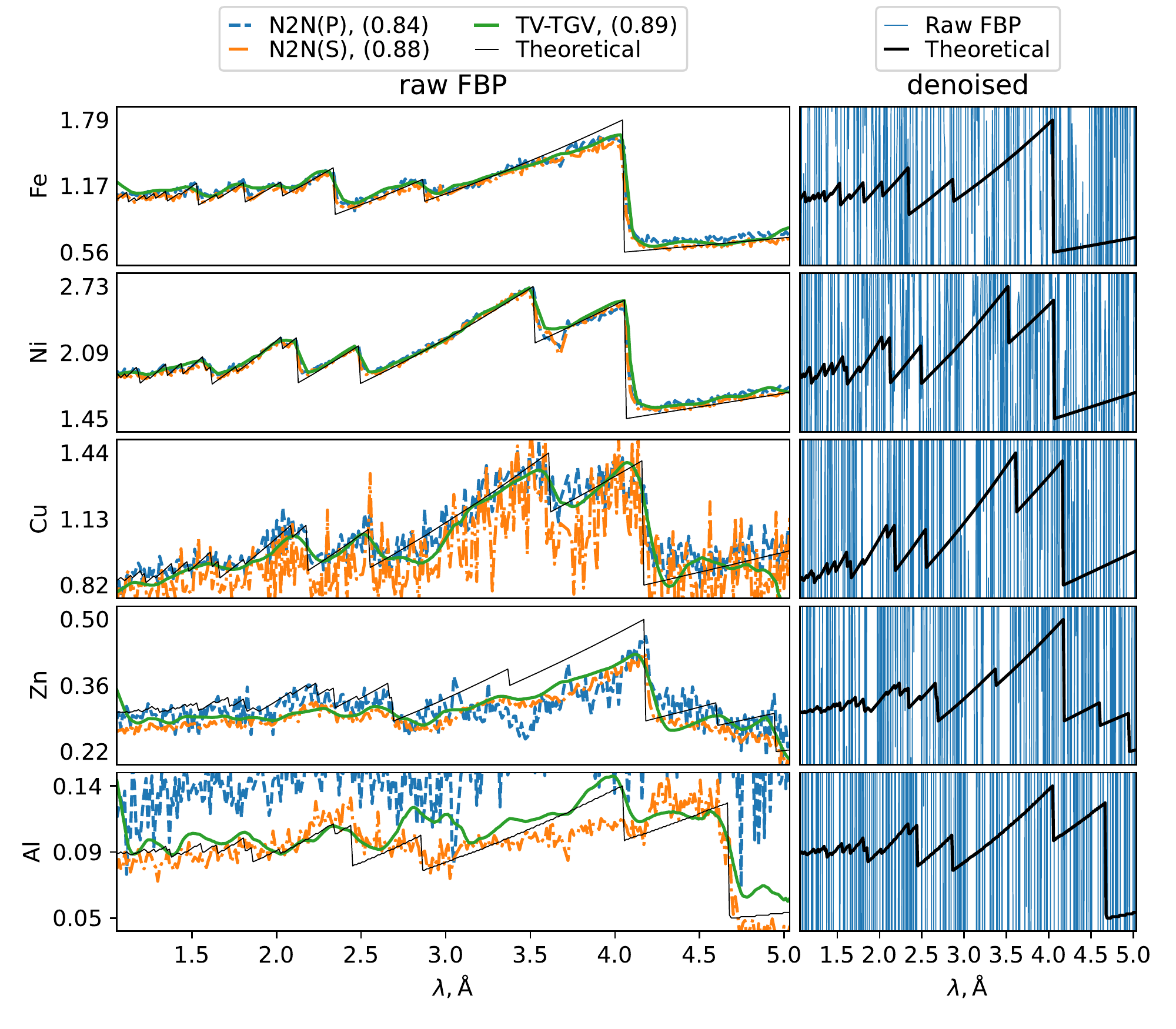}
\caption{Qualitative comparison of denoising techniques in the spectral domain for the neutron imaging dataset. 
For each material, we selected one representative voxel, and present the theoretical and empirical spectra. Left: results of TV-TGV, N2N done on slices, and N2N done on projections are presented; right: the spectra before denoising are presented. All results are presented for the datatset with 339 energy bins. We note, that N2N provides sharper edges, but noisier predictions.}
\label{fig:neutron:profiles}
\end{figure}

We begin with a visual comparison of different denoising approaches in the spectral domain (Figure~\ref{fig:neutron:profiles}). We perform the comparison for the 339 channels image as the same binning was used for the case study in~\cite{ametova2021crystalline}. The theoretical predictions provide the ground truth for the comparison. As in the previous case study, without denoising the conventional FBP reconstruction results are uninterpretable. The N2N performance is comparable to a TV-TGV reconstruction. TV-TGV provides smoother spectra at a cost of spectral and spatial resolution loss. In contrast, N2N results appear sharper spatially but noisier spectrally for low-attenuative materials. Hence, we conclude that there is a certain threshold noise level N2N can handle efficiently.

\begin{figure}
    \begin{subfigure}{1\textwidth}
      \centering
      \includegraphics[width=\linewidth]{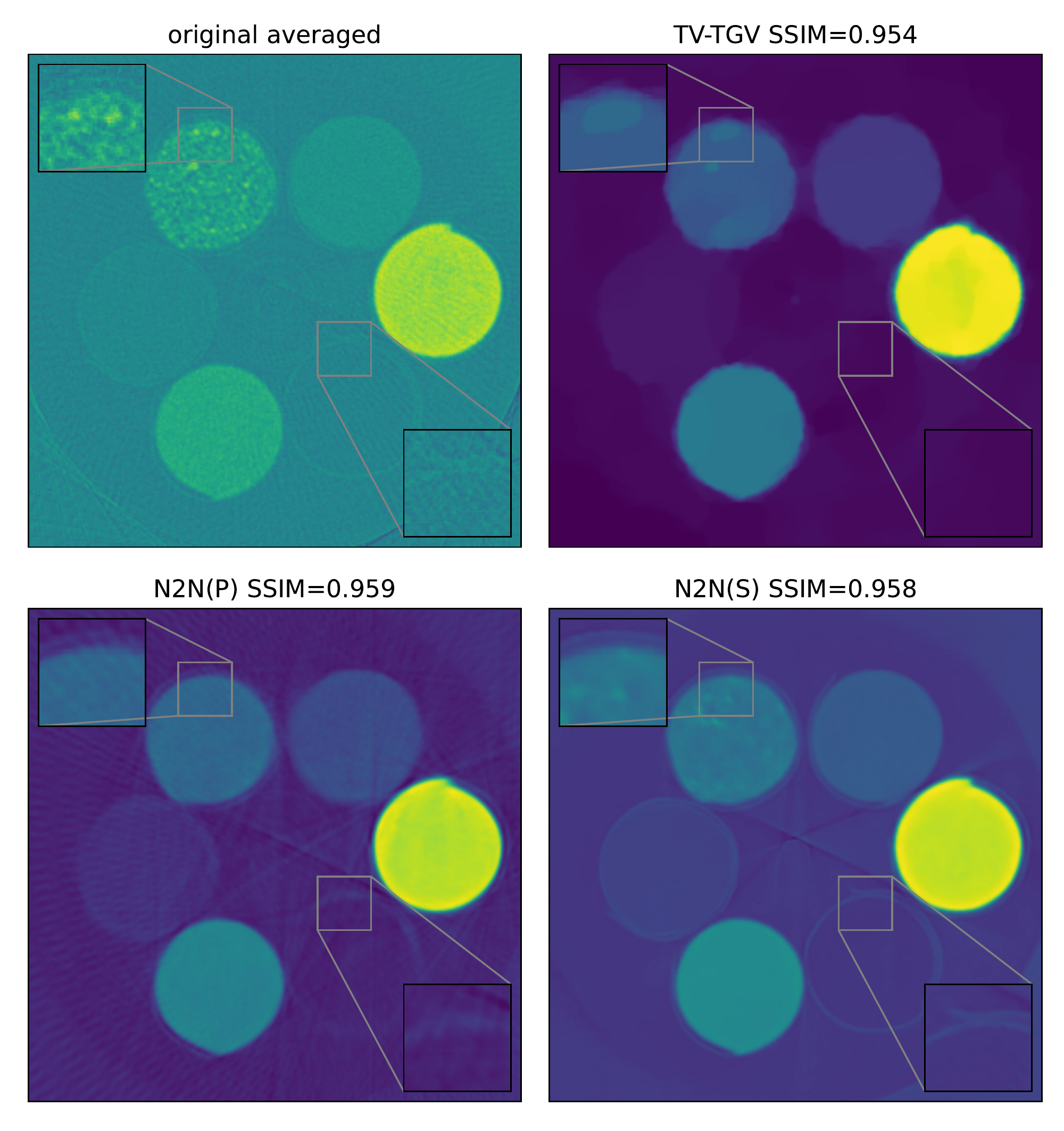}  
      \caption{Transverse slices showing white beam (sum of all energies) and single channel at the energy level of 1.63~\AA reconstruction for TV-TGV, N2N(S), and N2N(P).}
      \label{fig:neutron:slices}
    \end{subfigure}
    \newline
    \begin{subfigure}{\textwidth}
      \centering
        \begin{subfigure}{.49\textwidth}
          \centering
          \includegraphics[width=\linewidth]{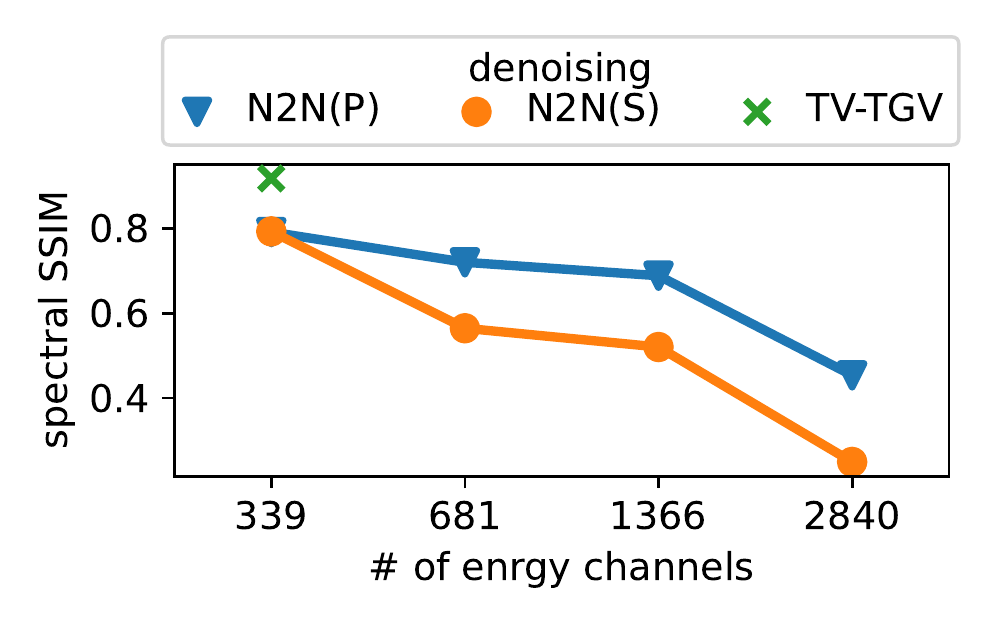}  
          \caption{SSIM between predicted and experimental spectra.}
          \label{fig:neutron:spectral_ssim}
        \end{subfigure}
        \begin{subfigure}{.49\textwidth}
          \centering
          \includegraphics[width=\linewidth]{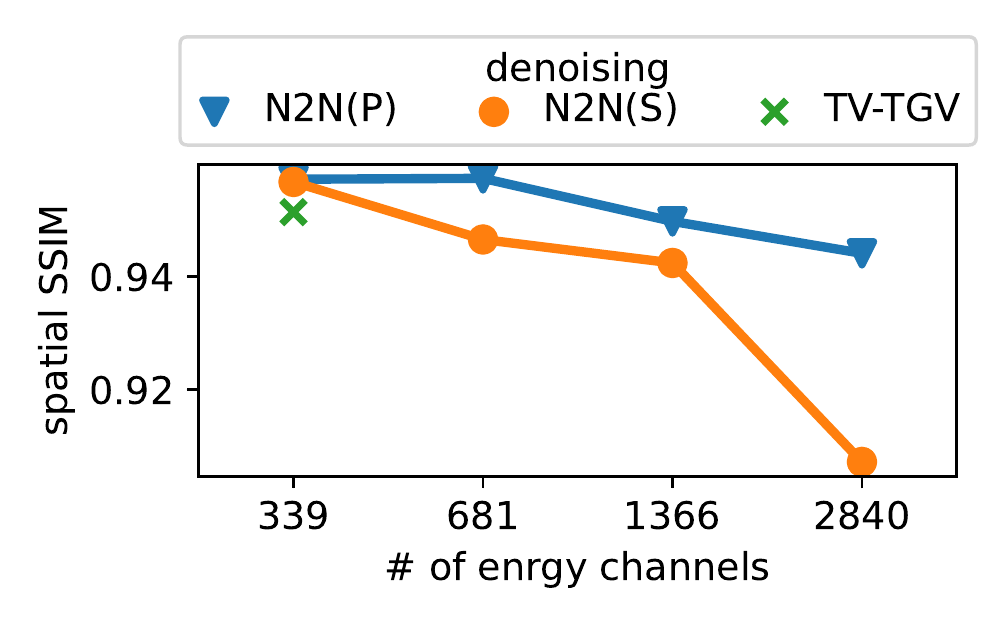}  
          \caption{Averaged SSIM between denoised slices and white beam slice reconstructed with FBP.}
          \label{fig:neutron:spatial_ssim}
        \end{subfigure}
    \end{subfigure}
    \caption{Qualitative (top) and quantitative (bottom) comparisons of the denoising methods for the neuron imaging are presented. For the quantitative comparison, we plot the dependency between the structural similarity index (separately in spectral and spatial domains) and the number of energy channels used. Since the change in the number of channels was done through the binning, the lower amount of channels corresponds to the lower amount of noise in the initial image. We note, that spatially Noise2Noise provides superior denoising.}
    \label{fig:neutron:quantitative}
\end{figure}

Figure~\ref{fig:neutron:slices} shows a comparison of the slices reconstructed from the white beam data (sum of all energies) and from data for a selected single energy channel for TV-TGV, N2N(S), and N2N(P), through direct comparisons of reconstructed slices in the transverse plane. While for Fe and Ni, both N2N(P) and N2N(S) perform comparably, for Cu and Al their performance differs. The attenuation of Al is drastically lower than other materials, which could lead to inconsistent predictions of the model for the projections when another material occludes the Al cylinder. This problem is not relevant for N2N(S). The Cu powder has a larger mean particle size than other powders (the mean particle size is comparable to the voxel size), hence, stronger spatial structures are visible in the cross-section. The structure changes randomly along the sample height. Therefore the N2N(S) model has less information about the structure and might fail to recover it correctly.

As a reference revealing structures, we use an FBP slice averaged across all energy levels, sacrificing spectral information for spatial. We also report the structural similarity index (SSIM) between the single-energy slices and the reference slice~\cite{Wang2004}. Both N2N approaches provide a sharper, more detailed image than TV-TGV. Interestingly, while N2N(S) provides a visually better, sharper image, this image has lower SSIM, compared to the N2N(P). We hypothesize, that this is caused by the unintentional reduction of the streak artifacts (highlighted in the top left callout in the N2N(P) slice). Streak artifacts are very common in tomographic imaging and are caused by insufficient angular sampling~\cite{kak2001principles}.

We next explore denoising quality in the spatial (Figure~\ref{fig:neutron:spatial_ssim}) and the spectral domain (Figure~\ref{fig:neutron:spectral_ssim}) given the increase of noise levels in the input data. We control noise levels by changing binning: the smaller is the binning step--the lower is SNR. We use the white beam slice reconstructed with FBP and the theoretical predictions for SSIM calculations in the spatial and spectral domains, respectively. While iterative reconstruction provides the best results for the spectral domain, it provides the worst result for the spatial domain. Excellent TV-TGV performance heavily capitalizes on the fact that the cylinders are homogeneous inside. In terms of SSIM, N2N(P) outperforms N2N(S) because N2N(S) additionally minimizes streak artifacts due to the angular undersampling, hence, the discrepancy between the reference image and the denoised one grows. 

Another important observation is that N2N can be computed for the higher number of channels. The training time of the model stays almost the same, around 20 hours on average for the full volume, calculated on a $4 \times A5000$ machine. After the training, the model is capable of inferencing one projection/slice at the rate of 20-30 energy channels per second. While TV-TGV reconstruction for one slice with 339 channels takes several hours to complete and reconstruction time increases with the increase in the number of channels or the number of slices.

\subsection{\emph{In Vivo} Cine-Radiography}
\label{sec:wasp}

The third case study considers a N2N application to cine-radiography. Cine-radiography or digital real-time radioscopy alias fluoroscopy are different realizations of time-resolved X-ray imaging techniques relying on X-ray projection imaging to study morphological evolution during technological or biological processes. In particular, for \textit{in vivo} or other dose-sensitive applications, the applicable dose and the detection efficiency of the imaging system limits acquisition times, constraining the total observation time or achievable SNR.

For this case study, we employed propagations-based phase contrast imaging (PB-PCI), which is particularly well suited for X-ray imaging of very weakly absorbing soft tissue in biological specimens in the sub-micron up to a few µm resolution range ~\cite{Fitzgerald2000}. 
The X-ray wavefield experiences a locally varying phase shift when traversing the specimen, which turns into measurable intensity contrast as a result of free-space wavefield propagation.
The object information can be reconstructed from the detected image interference pattern by algorithmic treatments (so-called \emph{phase retrieval} or PR for short~\cite{Lohse2020}). Here, we applied a convolution with a dedicated low-pass filter in the spatial domain. This so-called Paganin  filter~\cite{Paganin2002} heavily affects the noise distribution. On one hand, it significantly reduces high-frequency noise, hence, increases the Peak Signal-to-Noise Ratio (PSNR). On the other hand, low-frequency noise becomes more prominent causing so-called ``cloudy'' artifacts~\cite{paganin2004quantitative}. For a single image, the effect of low-frequency noise might be less disturbing. However, in a time-resolved cine-radiographic sequence, this effect leads to a highly disturbing flickering, since the position of these ``clouds'' changes randomly from frame to frame, which affects the interpretability of the images by experts.

\subsubsection{Data}

In this case study, we used a batch of \emph{in vivo} cine-radiographic data from a behavioral study visualizing the morphodynamics of parasitoid chalcid wasps emerging from their host eggs \cite{Spicker2023}. The full dataset contained $138$ videos, imaged with $15$~fps ($0.066$~s exposure time per frame) with lengths between $81$ and $7142$ frames per image series. The total number of frames is $263,875$.

We identified a sequence of 100 frames, where the wasp was completely still. From this, we calculated an average frame and used it as a low-noise reference image. This averaged image was used for qualitative results calculations. The average PSNR value before phase retrieval is $25.2$ with a standard deviation of $0.02$. After the Paganin phase retrieval, the PSNR increases to $35.9$ with a standard deviation of  $1.2$.

\subsubsection{Training and Analysis Details}

Because of the high dynamics in the sample's motions, we cannot use more than one frame as model input at one pass. We train the model $f_{\theta}$ by optimizing the loss 

\begin{equation}
    \mathbb{E}_{i,j} \Vert f_{\theta}(x_{i,j-1}) - x_{i,j} \Vert_1 \xrightarrow[\theta]{} \min,
\end{equation}

\noindent where $x_{i,j}$ stands for the frame number $j$ from the image sequence number $i$. We randomly divided all frames into training and validation sets in the $80/20$ ratio according to the index $i$. In addition, we noticed that in some cases the temporal resolution was not high enough to smoothly capture fast movements because the structure positions changed significantly between adjacent frames. We introduced additional filtering to alleviate potential blur caused by the large morphodynamical changes between neighboring frames. During the training, we discard the image pairs whose SSIM was below a manually optimized threshold.

\subsubsection{Results}

We applied the N2N denoising once before and once after phase retrieval.
Table~\ref{tab:wasp} summarizes PSNR and SSIM for both cases (the average of 100 frames without motion was used as a reference for metrics calculation). Applying denoising before the phase retrieval results in significant improvement in PSNR and SSIM. The benefits are maintained even after phase retrieval.  

\begin{figure}
      \begin{subfigure}{1\textwidth}
      \centering
      \includegraphics[width=\linewidth]{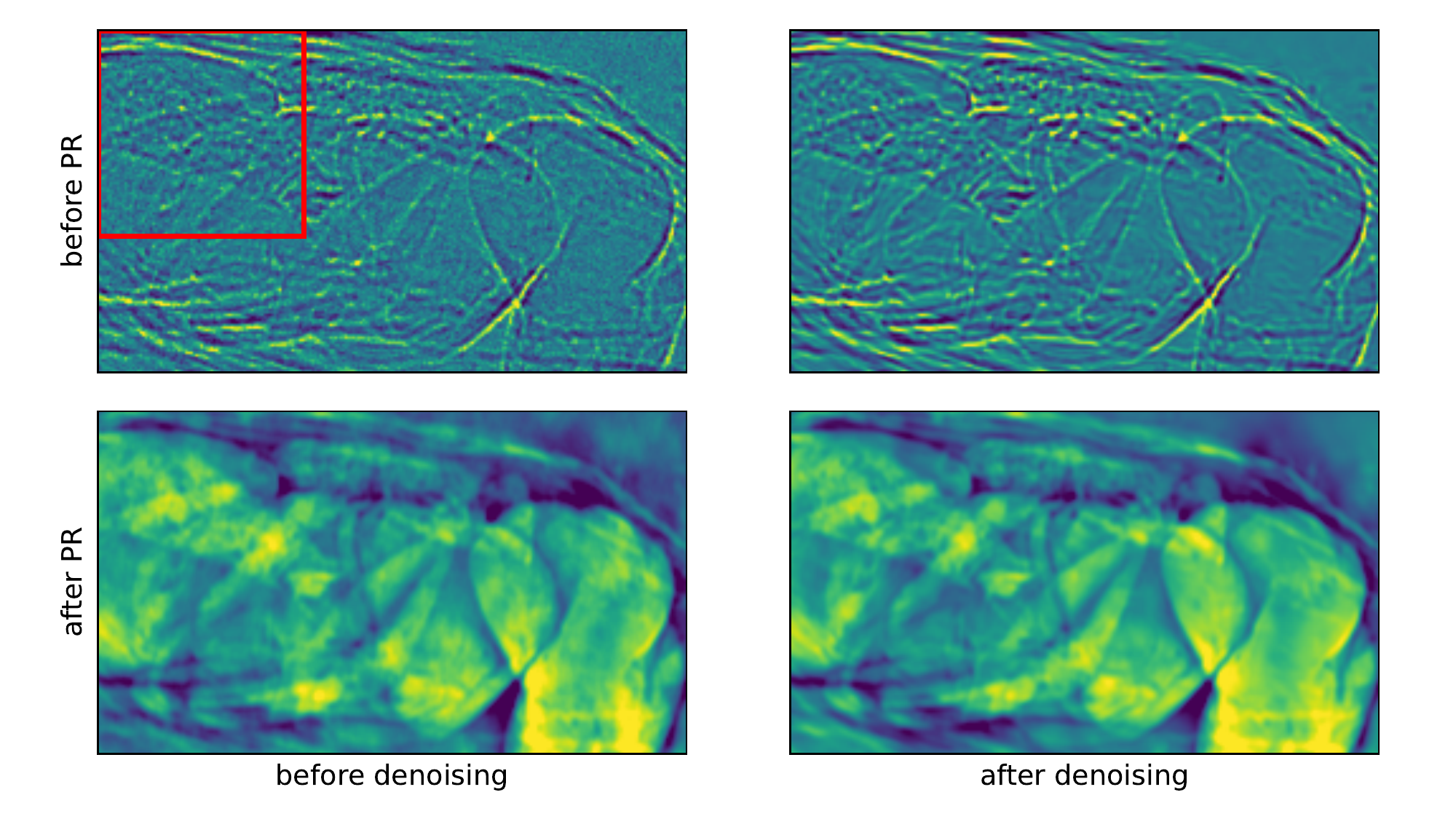}
      \caption{Overall view of the wasp.}
    \label{fig:wasp:overall}
    \end{subfigure}
    \newline
    \begin{subfigure}{\textwidth}
      \centering
        \begin{subfigure}[t]{.32\textwidth}
          \centering
          \includegraphics[width=\linewidth]{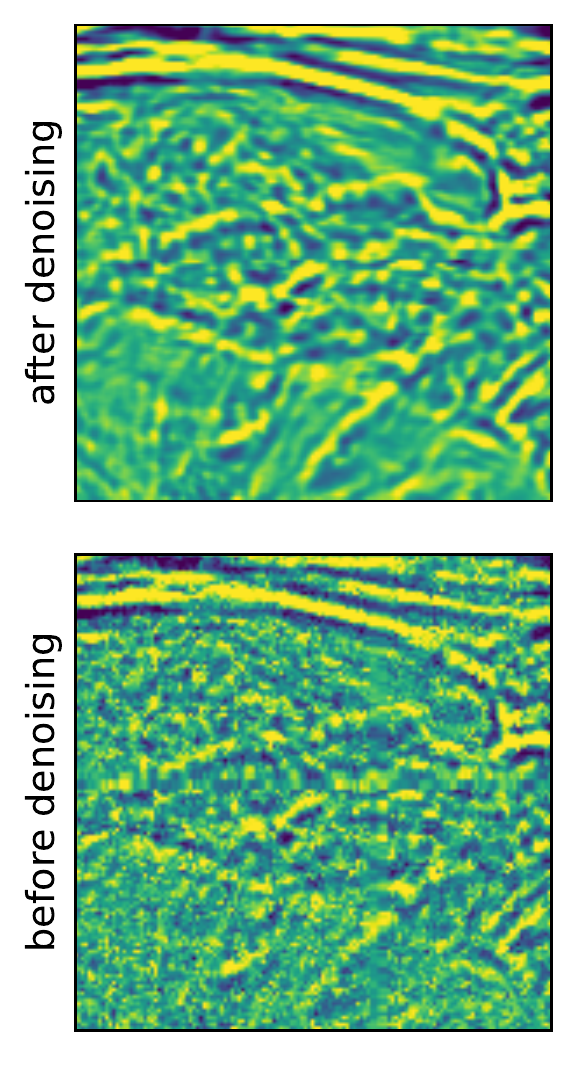}  
          \caption{Leg of the wasp before PR.}
          \label{fig:wasp:leg}
        \end{subfigure}
        \begin{subfigure}[t]{.32\textwidth}
          \centering
          \includegraphics[width=\linewidth]{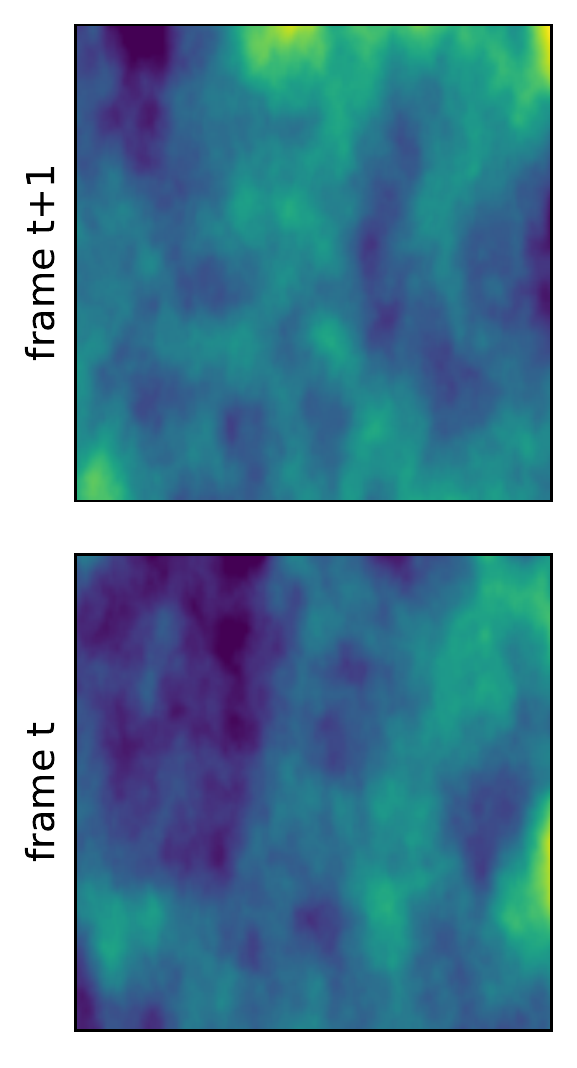}  
          \caption{Consequent frames before denoising.}
          \label{fig:wasp:noisy_flick}
        \end{subfigure}
        \begin{subfigure}[t]{.32\textwidth}
          \centering
          \includegraphics[width=\linewidth]{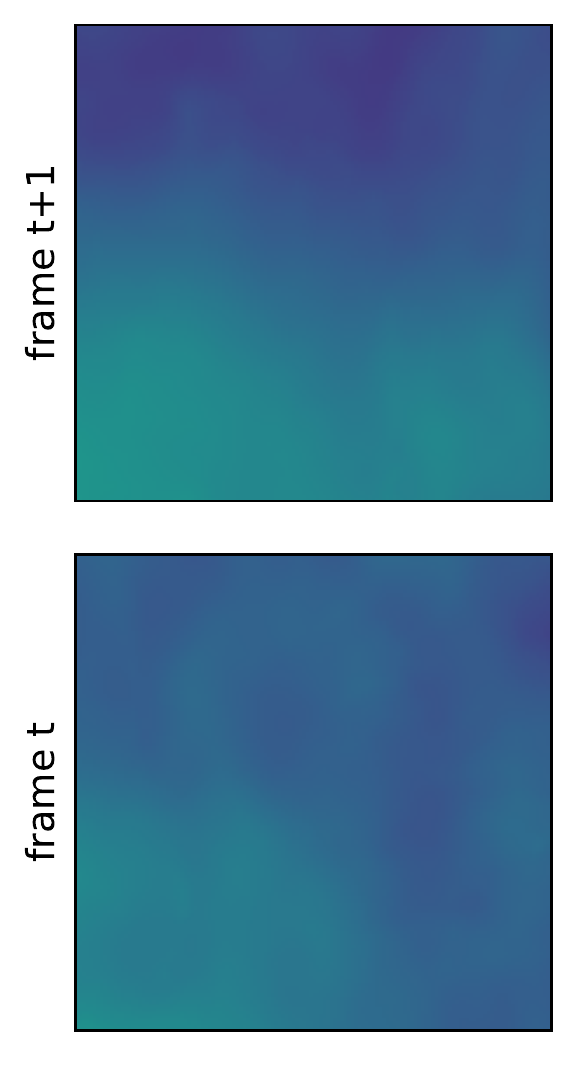}  
          \caption{Consequent frames after denoising.}
          \label{fig:wasp:denoised_flick}
        \end{subfigure}
    \end{subfigure}
    \caption{Qualitative examination of the denoising performed for the chalcid wasp cine-radiography. 
    We show the same cropped slice before and after the denoising. We also compare results before and after the phase retrieval. 
    Denoising is done only before phase retrieval. 
    In (b) we demonstrate an enlarged view of the wasp's leg before PR (a callout from the red rectangle in (a)). 
    In (c) and (d) we show consequent frames of the noise without the sample, before and after denoising. All four frames are plotted with the same value range. This demonstrates that not only the noise becomes less prominent, but also the evolution of the cloudy noise becomes less drastic after the denoising.}
      \label{fig:wasp:samples}
\end{figure}

To qualitatively assess the benefits of denoising done before the phase retrieval, we show exemplary frames in \Cref{fig:wasp:samples}.
Note that after the denoising and before phase retrieval the complex structures of the insect leg and interference fringes become more visible (\Cref{fig:wasp:leg}).
We also visually compare how the noise changes between consequent frames without (\Cref{fig:wasp:noisy_flick}) and with (\Cref{fig:wasp:denoised_flick}) denoising.
We note that the noise not only becomes less sharp without blurring the sample (\Cref{fig:wasp:overall}) but also produces less sudden changes in consequent frames.
This makes it easier to evaluate the morphodynamics or, reversely, would allow reducing the dose even further. While denoising made the images smoother, there is no drastic blur, and even relatively small details (\emph{e.g.}, legs or antennae) are preserved.

\begin{table}[]
    \centering
        \begin{tabular}{r||c|c|c|c}
             & \multicolumn{2}{c|}{measured before PR} & \multicolumn{2}{c}{measured after PR} \\
             & PSNR & SSIM & PSNR & SSIM \\
             \hline
             \hline
                no denoising & $25.2 \pm 4\times10^{-3}$ & $0.41 \pm 1\times10^{-4}$ & $36.0 \pm 0.2$ & $0.97 \pm 2\times10^{-3}$ \\
             \hline
                denoising before PR & $33.1 \pm 13\times10^{-3}$ & $0.49 \pm 3 \times 10^{-4}$ & $37.3 \pm 0.3$ & $0.98 \pm 1 \times 10^{-3}$ \\
             \hline
                denoising after PR & - & - & $36.0 \pm 0.3$ & $0.97 \pm 1 \times 10^{-3}$ \\
        \end{tabular}
    \caption{\label{tab:wasp} Quantitative comparison of the denoising done before and after phase retrieval for the chalcid wasp cine-radiography. We averaged 100 motion-free frames to use as the reference (noise-free) image for these calculations. We report mean values and 95\% confidence intervals. The denoising before phase retrieval provides a slight improvement in measures both before and after phase retrieval. }
\end{table}

\section{Discussion}
In this paper, we proposed and tested a way to relax the data constraints of the N2N method.
We traded the degree of similarity required from signals of individual image pairs for the number with structurally close image pairs. That is, instead of taking one pair with an identical signal, application to multi-channel imaging allows us to benefit from drawing tens or hundreds of pairs with close but not identical signals.
Through our experiments with different imaging modalities, we demonstrated the method's capability to significantly enhance image quality without over-smoothing along the energy or time domain. We also highlighted the vulnerability of the method to significantly dissimilar training image pairs (\Cref{sec:wasp}). We suggested an approach to overcoming this issue by discarding images based on some image similarity metrics.

In general, the method does not compensate for systematic artifacts present in all channels (energy bins or time frames), hence relevant corrections (such as centre-of-rotation compensation in CT) remain essential. On the other hand, we noticed that N2N applied to reconstructed slices significantly reduced the appearance of the ring artifacts (Figure \ref{fig:simulation_qualitative:140}) and the undersampling artifacts (Figure \ref{fig:neutron:slices}). The magnitude and the limits of this secondary effect are in the scope of future research.

We yet observed superficial ``cloudy'' artifacts present in homogeneous image regions (such as the background) after denoising. 
These artifacts do not have any significant effect on CT data as they have lower contrast than actual image features. 
However, in cine-radiographic time series, they affect the overall image perception as their location changes randomly from frame to frame introducing strongly disturbing flickering without denoising. The reduced flickering after denoising improves image interpretability as it has very low contrast compared to image features and becomes more of a cosmetic effect as the human eye is still sensitive to it. A way around this is to generate a few images for each frame by adding some Gaussian noise, i.e. to increase the noise level, and take a median value of the resulting denoised images. However, increasing the number of used denoised images or the variance of the noise leads to blurring. We set these parameters, guided by the expert judgment on the resulting image.

N2N assumes that the image pairs have equal signal values and independent noise drawn from the same distribution. Strictly speaking, both assumptions might be violated in spectral and time-resolved imaging where values in each individual channel are either energy or time-dependent, and noise distribution might be partially correlated (see data discussion in \ref{sec:neutron:data} for details). The application of the method to such data is based on the assumption that the variability of noise between the twin images is larger than the variability of the signal.

In spectral CT, N2N can be applied to both projection images and to tomographic slices after reconstruction. Any corrections in the projection domain are challenging as they might cause or exaggerate existing inconsistency between projections (a consistent sinogram has strong restrictions expressed as Helgason–Ludwig consistency condition~\cite{helgason1965radon}). However, our empirical studies did not show any noticeable artifacts due to this inconsistency.

\section{Conclusion}

In this paper, we explored the applicability of the N2N method to the denoising of time or energy-resolved radiographic image sequences and related 4D tomographic reconstructions. N2N is a distribution-agnostic method, it does not explicitly assume any particular noise or signal properties. The only requirement originally proposed was the ability to sample pairs of images that share a common signal but have independent and identically distributed noise. In this paper, we have demonstrated that this requirement, while not exactly met by the multichannel data, can be relaxed to successfully apply the method. 

The presented case studies showed that this method offers a robust and efficient alternative to conventional denoising methods and regularized iterative reconstruction methods. The N2N method does not require fine-tuning of parameters or handcrafted regularization terms for any new dataset. Therefore, its application can be heavily automated. Finally, the N2N method relies on a rather intuitive assumption, hence it can be easily explained to non-experts in the ML domain and smoothly introduced into their measurement practice.

\begin{backmatter}
\bmsection{Funding} We gratefully acknowledge beamtime RB1820541 at the IMAT Beamline of the ISIS Neutron and Muon Source, Harwell, UK. The research was supported by the HIGH-LIFE and SMART-MORPH projects of the Federal Ministry of Education and Research (BMBF) and by the Baden-Württemberg Ministry of Science as part of the Excellence Strategy of the German Federal and State Governments.

\bmsection{Acknowledgments} This work made use of computational support by CoSeC, the Computational Science Centre for Research Communities, through the Collaborative Computational Project in Tomographic Imaging (CCPi).

\bmsection{Data Availability Statement} The neutron tomography dataset is openly available at the following URL/DOI: (\href{https://doi.org/10.5286/ISIS.E.RB1820541}{10.5286/ISIS.E.RB1820541}).
The code to reproduce the phantom and the dataset will be made publicly available upon the publication of the paper.

\end{backmatter}

\bibliography{sample, references}

\begin{thebibliography}{10}
\newcommand{\enquote}[1]{``#1''}

\bibitem{Lehtinen2018}
J.~Lehtinen, J.~Munkberg, J.~Hasselgren, S.~Laine, T.~Karras, M.~Aittala, and
  T.~Aila, \enquote{{Noise2Noise: Learning image restoration without clean
  data},} {\protect\JournalTitle{35th International Conference on Machine
  Learning, ICML 2018}} \textbf{7}, 4620--4631 (2018).

\bibitem{Moosmann2013}
J.~Moosmann, A.~Ershov, V.~Altapova, T.~Baumbach, M.~S. Prasad, C.~LaBonne,
  X.~Xiao, J.~Kashef, and R.~Hofmann, \enquote{{X-ray phase-contrast in vivo
  microtomography probes new aspects of Xenopus gastrulation},}
  {\protect\JournalTitle{Nature}} \textbf{497}, 374--377 (2013).

\bibitem{warr2021enhanced}
R.~Warr, E.~Ametova, R.~J. Cernik, G.~Fardell, S.~Handschuh, J.~S.
  J{\o}rgensen, E.~Papoutsellis, E.~Pasca, and P.~J. Withers, \enquote{Enhanced
  hyperspectral tomography for bioimaging by spatiospectral reconstruction,}
  {\protect\JournalTitle{Scientific reports}} \textbf{11}, 1--13 (2021).

\bibitem{Ilesanmi2021}
A.~E. Ilesanmi and T.~O. Ilesanmi, \enquote{{Methods for image denoising using
  convolutional neural network: a review},} {\protect\JournalTitle{Complex {\&}
  Intelligent Systems}} \textbf{7}, 2179--2198 (2021).

\bibitem{Gonzalez2008}
R.~C. Gonzalez and R.~E. Woods, \emph{{Digital Image Processing}} (Prentice
  Hall, 2008).

\bibitem{Buades2005}
A.~Buades, B.~Coll, and J.~M. Morel, \enquote{{A non-local algorithm for image
  denoising},} in \emph{Proceedings - 2005 IEEE Computer Society Conference on
  Computer Vision and Pattern Recognition, CVPR 2005,}  vol.~II (IEEE, 2005),
  pp. 60--65.

\bibitem{Fan2019}
L.~Fan, F.~Zhang, H.~Fan, and C.~Zhang, \enquote{{Brief review of image
  denoising techniques},} {\protect\JournalTitle{Visual Computing for Industry,
  Biomedicine, and Art}} \textbf{2}, 7 (2019).

\bibitem{gu2019brief}
S.~Gu and R.~Timofte, \enquote{A brief review of image denoising algorithms and
  beyond,} {\protect\JournalTitle{Inpainting and Denoising Challenges}} pp.
  1--21 (2019).

\bibitem{rodriguez2013total}
P.~Rodr{\'\i}guez, \enquote{Total variation regularization algorithms for
  images corrupted with different noise models: a review,}
  {\protect\JournalTitle{Journal of Electrical and Computer Engineering}}
  \textbf{2013} (2013).

\bibitem{Batson2019}
J.~Batson and L.~Royer, \enquote{{Noise2Seif: Blind denoising by
  self-supervision},} {\protect\JournalTitle{36th International Conference on
  Machine Learning, ICML 2019}} \textbf{2019-June}, 826--835 (2019).

\bibitem{Krull2019}
A.~Krull, T.~O. Buchholz, and F.~Jug, \enquote{{Noise2void-Learning denoising
  from single noisy images},} {\protect\JournalTitle{Proceedings of the IEEE
  Computer Society Conference on Computer Vision and Pattern Recognition}}
  \textbf{2019-June}, 2124--2132 (2019).

\bibitem{Papkov2021}
M.~Papkov, K.~Roberts, L.~A. Madissoon, J.~Shilts, O.~Bayraktar, D.~Fishman,
  K.~Palo, and L.~Parts, \enquote{{Noise2Stack: Improving Image Restoration by
  Learning from Volumetric Data},} {\protect\JournalTitle{Lecture Notes in
  Computer Science (including subseries Lecture Notes in Artificial
  Intelligence and Lecture Notes in Bioinformatics)}} \textbf{12964 LNCS},
  99--108 (2021).

\bibitem{Vincent2008}
P.~Vincent, H.~Larochelle, Y.~Bengio, and P.~A. Manzagol, \enquote{{Extracting
  and composing robust features with denoising autoencoders},} in
  \emph{Proceedings of the 25th International Conference on Machine Learning,}
  (ACM Press, New York, New York, USA, 2008), pp. 1096--1103.

\bibitem{Prakash2022}
M.~Prakash, M.~Delbracio, P.~Milanfar, and F.~Jug, \enquote{{Interpretable
  Unsupervised Diversity Denoising and Artefact Removal},}
  {\protect\JournalTitle{International Conference on Learning Representations}}
   (2022).

\bibitem{Dalsasso2022}
E.~Dalsasso, L.~Denis, and F.~Tupin, \enquote{{As if by Magic: Self-Supervised
  Training of Deep Despeckling Networks with MERLIN},}
  {\protect\JournalTitle{IEEE Transactions on Geoscience and Remote Sensing}}
  \textbf{60} (2022).

\bibitem{PavelIakubovskii2019}
{Pavel Iakubovskii}, \enquote{{Segmentation Models Pytorch},}  (2019).

\bibitem{davis2008modelling}
G.~Davis, N.~Jain, and J.~Elliott, \enquote{A modelling approach to beam
  hardening correction,} in \emph{Developments in X-ray Tomography VI,}  vol.
  7078 (SPIE, 2008), pp. 423--432.

\bibitem{egan20153d}
C.~Egan, S.~Jacques, M.~Wilson, M.~Veale, P.~Seller, A.~Beale, R.~Pattrick,
  P.~Withers, and R.~Cernik, \enquote{3d chemical imaging in the laboratory by
  hyperspectral x-ray computed tomography,} {\protect\JournalTitle{Scientific
  reports}} \textbf{5}, 1--9 (2015).

\bibitem{Pedregosa2012}
F.~Pedregosa, G.~Varoquaux, A.~Gramfort, V.~Michel, B.~Thirion, O.~Grisel,
  M.~Blondel, A.~M{\"{u}}ller, J.~Nothman, G.~Louppe, P.~Prettenhofer,
  R.~Weiss, V.~Dubourg, J.~Vanderplas, A.~Passos, D.~Cournapeau, M.~Brucher,
  M.~Perrot, and E.~Duchesnay, \enquote{{Scikit-learn: Machine Learning in
  Python},} {\protect\JournalTitle{Journal of Machine Learning Research}}
  \textbf{12}, 2825--2830 (2011).

\bibitem{getzin2018increased}
M.~Getzin, J.~J. Garfield, D.~S. Rundle, U.~Kruger, A.~P. Butler, M.~Gkikas,
  and G.~Wang, \enquote{Increased separability of k-edge nanoparticles by
  photon-counting detectors for spectral micro-ct,}
  {\protect\JournalTitle{Journal of X-ray science and technology}} \textbf{26},
  707--726 (2018).

\bibitem{tuszynski2006}
J.~Tuszynski, \enquote{2022 photonattenuation--software for modeling of photons
  passing through different materials,}  (2006).

\bibitem{punnoose2016spektr}
J.~Punnoose, J.~Xu, A.~Sisniega, W.~Zbijewski, and J.~Siewerdsen,
  \enquote{spektr 3.0—a computational tool for x-ray spectrum modeling and
  analysis,} {\protect\JournalTitle{Medical physics}} \textbf{43}, 4711--4717
  (2016).

\bibitem{jorgensen2021core}
J.~S. J{\o}rgensen, E.~Ametova, G.~Burca, G.~Fardell, E.~Papoutsellis,
  E.~Pasca, K.~Thielemans, M.~Turner, R.~Warr, W.~R. Lionheart \emph{et~al.},
  \enquote{{Core Imaging Library} -- {Part I}: a versatile {Python} framework
  for tomographic imaging,} {\protect\JournalTitle{Philosophical Transactions
  of the Royal Society A}} \textbf{379}, 20200192 (2021).

\bibitem{ametova2021crystalline}
E.~Ametova, G.~Burca, S.~Chilingaryan, G.~Fardell, J.~S. J{\o}rgensen,
  E.~Papoutsellis, E.~Pasca, R.~Warr, M.~Turner, W.~R. Lionheart \emph{et~al.},
  \enquote{Crystalline phase discriminating neutron tomography using advanced
  reconstruction methods,} {\protect\JournalTitle{Journal of Physics D: Applied
  Physics}} \textbf{54}, 325502 (2021).

\bibitem{fundamentals1993nuclear}
D.~Fundamentals, \enquote{Nuclear physics and reactor theory,} Tech. rep.,
  Technical Report (1993).

\bibitem{Santisteban2001}
J.~R. Santisteban, L.~Edwards, A.~Steuwer, and P.~J. Withers,
  \enquote{{Time-of-flight neutron transmission diffraction},}
  {\protect\JournalTitle{Journal of Applied Crystallography}} \textbf{34},
  289--297 (2001).

\bibitem{kockelmann2007energy}
W.~Kockelmann, G.~Frei, E.~H. Lehmann, P.~Vontobel, and J.~R. Santisteban,
  \enquote{Energy-selective neutron transmission imaging at a pulsed source,}
  {\protect\JournalTitle{Nuclear Instruments and Methods in Physics Research
  Section A: Accelerators, Spectrometers, Detectors and Associated Equipment}}
  \textbf{578}, 421--434 (2007).

\bibitem{santisteban2002engineering}
J.~R. Santisteban, L.~Edwards, M.~E. Fizpatrick, A.~Steuwer, and P.~J. Withers,
  \enquote{Engineering applications of {Bragg}-edge neutron transmission,}
  {\protect\JournalTitle{Applied Physics A}} \textbf{74}, s1433--s1436 (2002).

\bibitem{strobl2009advances}
M.~Strobl, I.~Manke, N.~Kardjilov, A.~Hilger, M.~Dawson, and J.~Banhart,
  \enquote{Advances in neutron radiography and tomography,}
  {\protect\JournalTitle{Journal of Physics D: Applied Physics}} \textbf{42},
  243001 (2009).

\bibitem{Bentley2020}
P.~M. Bentley, \enquote{{Instrument suite cost optimisation in a science
  megaproject},} {\protect\JournalTitle{Journal of Physics Communications}}
  \textbf{4}, 045014 (2020).

\bibitem{jorgensen2019neutron}
J.~J{\o}rgensen, E.~Ametova, G.~Burca, G.~Fardell, E.~Papoutsellis, E.~Pasca,
  A.~Liptak, D.~Kazantsev, W.~Lionheart, and M.~Turner, \enquote{Neutron {TOF}
  imaging phantom data to quantify hyperspectral reconstruction algorithms,}
  {\protect\JournalTitle{STFC ISIS Neutron and MuonSource}}  (2019).

\bibitem{burca2013modelling}
G.~Burca, W.~Kockelmann, J.~James, and M.~E. Fitzpatrick, \enquote{Modelling of
  an imaging beamline at the {ISIS} pulsed neutron source,}
  {\protect\JournalTitle{Journal of Instrumentation}} \textbf{8}, P10001
  (2013).

\bibitem{kockelmann2018time}
W.~Kockelmann, T.~Minniti, D.~E. Pooley, G.~Burca, R.~Ramadhan, F.~A. Akeroyd,
  G.~D. Howells, C.~Moreton-Smith, D.~P. Keymer, J.~Kelleher, S.~Kabra, T.~L.
  Lee, R.~Ziesche, A.~Reid, G.~Vitucci, G.~Gorini, D.~Micieli, R.~G. Agostino,
  V.~Formoso, F.~Aliotta, R.~Ponterio, S.~Trusso, G.~Salvato, C.~Vasi,
  F.~Grazzi, K.~Watanabe, J.~W.~L. Lee, A.~S. Tremsin, J.~B. McPhate, D.~Nixon,
  N.~Draper, W.~Halcrow, and J.~Nightingale, \enquote{Time-of-flight neutron
  imaging on {IMAT@ISIS}: A new user facility for materials science,}
  {\protect\JournalTitle{Journal of Imaging}} \textbf{4} (2018).

\bibitem{tremsin2012high}
A.~S. Tremsin, J.~V. Vallerga, J.~B. McPhate, O.~H. Siegmund, and R.~Raffanti,
  \enquote{High resolution photon counting with mcp-timepix quad parallel
  readout operating at > 1 khz frame rates,} {\protect\JournalTitle{IEEE
  transactions on nuclear science}} \textbf{60}, 578--585 (2012).

\bibitem{rudin1992nonlinear}
L.~I. Rudin, S.~Osher, and E.~Fatemi, \enquote{Nonlinear total variation based
  noise removal algorithms,} {\protect\JournalTitle{Physica D: nonlinear
  phenomena}} \textbf{60}, 259--268 (1992).

\bibitem{sidky2006accurate}
E.~Y. Sidky, C.-M. Kao, and X.~Pan, \enquote{Accurate image reconstruction from
  few-views and limited-angle data in divergent-beam {CT},}
  {\protect\JournalTitle{Journal of X-ray Science and Technology}} \textbf{14},
  119--139 (2006).

\bibitem{boin2012nxs}
M.~Boin, \enquote{{NXS}: a program library for neutron cross section
  calculations,} {\protect\JournalTitle{Journal of Applied Crystallography}}
  \textbf{45}, 603--607 (2012).

\bibitem{bredies2010total}
K.~Bredies, K.~Kunisch, and T.~Pock, \enquote{Total generalized variation,}
  {\protect\JournalTitle{SIAM Journal on Imaging Sciences}} \textbf{3},
  492--526 (2010).

\bibitem{Papoutsellis2021}
E.~Papoutsellis, E.~Ametova, C.~Delplancke, G.~Fardell, J.~S. J{\o}rgensen,
  E.~Pasca, M.~Turner, R.~Warr, W.~R.~B. Lionheart, and P.~J. Withers,
  \enquote{{Core Imaging Library - Part II: multichannel reconstruction for
  dynamic and spectral tomography},} {\protect\JournalTitle{Philosophical
  Transactions of the Royal Society A: Mathematical, Physical and Engineering
  Sciences}} \textbf{379}, 20200193 (2021).

\bibitem{evelina_ametova_2021_4884710}
E.~Ametova, G.~Burca, S.~Chilingaryan, G.~Fardell, J.~S. Jørgensen,
  E.~Papoutsellis, E.~Pasca, R.~Warr, M.~Turner, W.~R.~B. Lionheart, and P.~J.
  Withers, \enquote{{Code to reproduce results of "Crystalline phase
  discriminating neutron tomography using advanced reconstruction methods"},}
  (2021).

\bibitem{Wang2004}
Z.~Wang, A.~Bovik, H.~Sheikh, and E.~Simoncelli, \enquote{{Image Quality
  Assessment: From Error Visibility to Structural Similarity},}
  {\protect\JournalTitle{IEEE Transactions on Image Processing}} \textbf{13},
  600--612 (2004).

\bibitem{kak2001principles}
A.~C. Kak and M.~Slaney, \emph{Principles of computerized tomographic imaging}
  (SIAM, 2001).

\bibitem{Fitzgerald2000}
R.~Fitzgerald, \enquote{{Phase‐Sensitive X‐Ray Imaging},}
  {\protect\JournalTitle{Physics Today}} \textbf{53}, 23--26 (2000).

\bibitem{Lohse2020}
L.~M. Lohse, A.~L. Robisch, M.~T{\"{o}}pperwien, S.~Maretzke, M.~Krenkel,
  J.~Hagemann, and T.~Salditt, \enquote{{A phase-retrieval toolbox for X-ray
  holography and tomography},} {\protect\JournalTitle{Journal of Synchrotron
  Radiation}} \textbf{27}, 852--859 (2020).

\bibitem{Paganin2002}
D.~Paganin, S.~C. Mayo, T.~E. Gureyev, P.~R. Miller, and S.~W. Wilkins,
  \enquote{{Simultaneous phase and amplitude extraction from a single defocused
  image of a homogeneous object},} {\protect\JournalTitle{Journal of
  Microscopy}} \textbf{206}, 33--40 (2002).

\bibitem{paganin2004quantitative}
D.~Paganin, A.~Barty, P.~McMahon, and K.~A. Nugent, \enquote{Quantitative
  phase-amplitude microscopy. iii. the effects of noise,}
  {\protect\JournalTitle{Journal of microscopy}} \textbf{214}, 51--61 (2004).

\bibitem{Spicker2023}
R.~Spiecker, P.~Pfeiffer, A.~Biswal, M.~Shcherbinin, M.~Spiecker,
  H.~Hessdorfer, M.~Hurst, Y.~Zharov, V.~Bellucci, T.~Farag{\'{o}}, M.~Zuber,
  A.~Herz, A.~Cecilia, M.~Czyzycki, D.~Novikov, C.~S.~B. Dias, L.~Krogmann,
  E.~Hamann, T.~van~de Kamp, and T.~Baumbach, \enquote{{Bragg magnifier based
  dose-efficient in vivo X-ray imaging at micrometer resolution},}
  {\protect\JournalTitle{submitted}}  (2023).

\bibitem{helgason1965radon}
S.~Helgason, \enquote{The radon transform on euclidean spaces, compact
  two-point homogeneous spaces and grassmann manifolds,}
  {\protect\JournalTitle{Acta Mathematica}} \textbf{113}, 153--180 (1965).

\end{thebibliography}

\end{document}